\documentclass[pdflatex,sn-mathphys-num]{sn-jnl}

\usepackage{graphicx}%
\usepackage{multirow}%
\usepackage{amsmath,amssymb,amsfonts}%
\usepackage{amsthm}%
\usepackage{mathrsfs}%
\usepackage[title]{appendix}%
\usepackage{xcolor}%
\usepackage{textcomp}%
\usepackage{manyfoot}%
\usepackage{booktabs}%
\usepackage{algorithm}%
\usepackage{algorithmicx}%
\usepackage{algpseudocode}%
\usepackage{listings}%

\usepackage{mathtools}
\usepackage{placeins}
\usepackage{xspace}
\usepackage{siunitx}
\usepackage{makecell}
\usepackage{array}

\newcommand{\methodname}{GLENS\xspace}
\newcommand{\neighbormodel}{\texttt{Neighborhood Structure Model}\xspace}
\newcommand{\solvermodel}{\texttt{Solver Behavior Model}\xspace}

\theoremstyle{thmstyleone}%
\theoremstyle{thmstyletwo}%

\theoremstyle{thmstylethree}%
\newtheorem{definition}{Definition}%

\begin{document}

\title[GLENS]{
\methodname: Global Search via Learning from Solver Iterates with Diffusion Models
}

\author*[1]{\fnm{Anjian} \sur{Li}}\email{anjianl@princeton.edu}

\author[2]{\fnm{Bartolomeo} \sur{Stellato}}\email{bstellato@princeton.edu}
\author[3]{\fnm{Ryne} \sur{Beeson}}\email{ryne@princeton.edu}

\affil*[1]{\orgdiv{Department of Electrical and Computer Engineering}, \orgname{Princeton University}, \orgaddress{\city{Princeton}, \postcode{08544}, \state{NJ}, \country{USA}}}

\affil[2]{\orgdiv{Department of Operations Research and Financial Engineering}, \orgname{Princeton University}, \orgaddress{\city{Princeton}, \postcode{08544}, \state{NJ}, \country{USA}}}

\affil[3]{\orgdiv{Department of Mechanical and Aerospace Engineering}, \orgname{Princeton University}, \orgaddress{\city{Princeton}, \postcode{08544}, \state{NJ}, \country{USA}}}

\abstract{
We consider the problem of generating a large collection of initial guesses for local minima of multimodal non-convex continuous optimization problems.
The goal is for these initial guesses to be high-quality (i.e., a numerical solver converges quickly) and diverse (i.e., represent many different local minima).
Identifying multiple locally optimal solutions enables flexible downstream decision-making, but typically requires expensive global search.
Existing data-driven methods predict initial guesses using only the final converged optima from offline solver runs, which discards information about the local neighborhoods of solutions and limits the available training data.
We propose GLENS (\textbf{G}lobal Search via \textbf{Le}ar\textbf{n}ing from \textbf{S}olver Iterates), a data-efficient global search method that leverages intermediate solver iterates as free data augmentation.
GLENS consists of two components: a neighborhood structure model that uses diffusion models to learn the local geometry around optima conditioned on problem parameters, and a solver behavior model that learns refinement directions to further guide samples towards nearby optima during diffusion sampling.
Experiments on modified non-convex benchmark problems and a two-robot obstacle-avoidance navigation problem show that GLENS generates high-quality initial guesses while preserving the multimodal distribution of diverse local optima.
The resulting initial guesses lead to faster solver convergence across different problem settings and solvers.
We also analyze how key hyperparameter choices affect the performance.
}


\maketitle

\section{Introduction}\label{sec:introduction}

Non-convex continuous optimization problems arising in real-world applications often exhibit multiple local optima.
Beyond seeking a single global optimum, it is often desirable to identify a diverse set of locally optimal solutions, enabling trade-offs among different objectives and requirements.
Ideally, this diverse set of solutions is also high-quality, meaning that their objective values are reasonably close to the global optimum, which is likely unknowable. 
For example, in autonomous robot navigation, multiple feasible trajectories may differ in safety, time efficiency, or comfort~\cite{dauner2024navsim}; in spaceflight trajectory design, qualitatively different trajectories may differ in time-of-flight, fuel consumption, or flyby sequences~\cite{englander2017automated}.
Access to such diverse candidate solutions provides greater flexibility for downstream decision-making.

However, finding a diverse collection of high-quality local optima for non-convex problems typically requires an expensive global search.
A widely used approach is the hybrid method, e.g.,\ multi-start~\cite{locatelli2016global} or Monotonic Basin Hopping (MBH)~\cite{leary2000global}, which combines global sampling with local search.
Multiple initial guesses are sampled from a heuristic distribution over the solution space, and a gradient-based numerical solver refines each initial guess through a sequence of iterates that converges to a nearby optimum~\cite{locatelli2016global}.
For high-dimensional and highly non-convex problems, identifying suitable initial guesses for the numerical solver is increasingly difficult, and the global search quickly becomes computationally expensive.

An alternative paradigm is amortized optimization~\cite{amos2023tutorial}, which solves many instances of a parametric optimization problem offline and learns how solutions~\cite{chen2022large} or solving strategies~\cite{cauligi2021coco,bertsimas2021voice} vary with problem parameters, so that high-quality solutions can be predicted efficiently for new instances at test time.
In this data-driven setting, multilayer perceptrons (MLPs) have been adopted for predicting the optimal solution of convex problems~\cite{JMLR:v25:23-1174}.
For problems with multiple local optima, generative models such as diffusion models~\cite{song2019generative,ho2020denoising} are used for sampling solution distributions, with applications to mixed-integer programming~\cite{sun2023difusco} and non-convex trajectory optimization~\cite{li2025diffusolve,graebner2025global}.

Despite their promise, learning-based methods for non-convex optimization face a significant challenge: data scarcity.
Generative models require large amounts of diverse, high-quality training data to learn the relationship between problem parameters and solution distributions~\cite{beeson2024global}.
Curating such datasets requires running expensive global searches across many problem instances, introducing a substantial computational bottleneck.

In addition, existing learning-based amortized optimization methods often collect only the final converged solutions produced by the numerical solvers~\cite{chen2022large,JMLR:v25:23-1174,li2025diffusolve}, while discarding the intermediate iterates generated during the solving process. 
Although this practice maintains solution quality, including feasibility and local optimality, it ignores valuable information about the neighborhood structure surrounding optimal solutions.
For example, prior work has shown that in trajectory optimization, locally optimal solutions can lie very close to infeasible regions~\cite{li2024constraint,li2025aligning}.
Without access to information about the solution's neighborhood structure, data-driven models may struggle to generalize robustly to variations in constraints or other problem parameters.

Motivated by these limitations, our key insight is to use intermediate iterates produced by numerical solvers as training data for learning-based amortized optimization.
Although these iterates are suboptimal, they encode rich information about the local neighborhood surrounding optimal solutions.
Importantly, solver-generated iterates provide free data augmentation, requiring no additional computational cost beyond the original optimization runs.
However, since intermediate iterates are suboptimal, directly incorporating them without proper modeling can degrade performance.

\begin{figure}[t]
    \centering
    \includegraphics[width=0.75\textwidth]{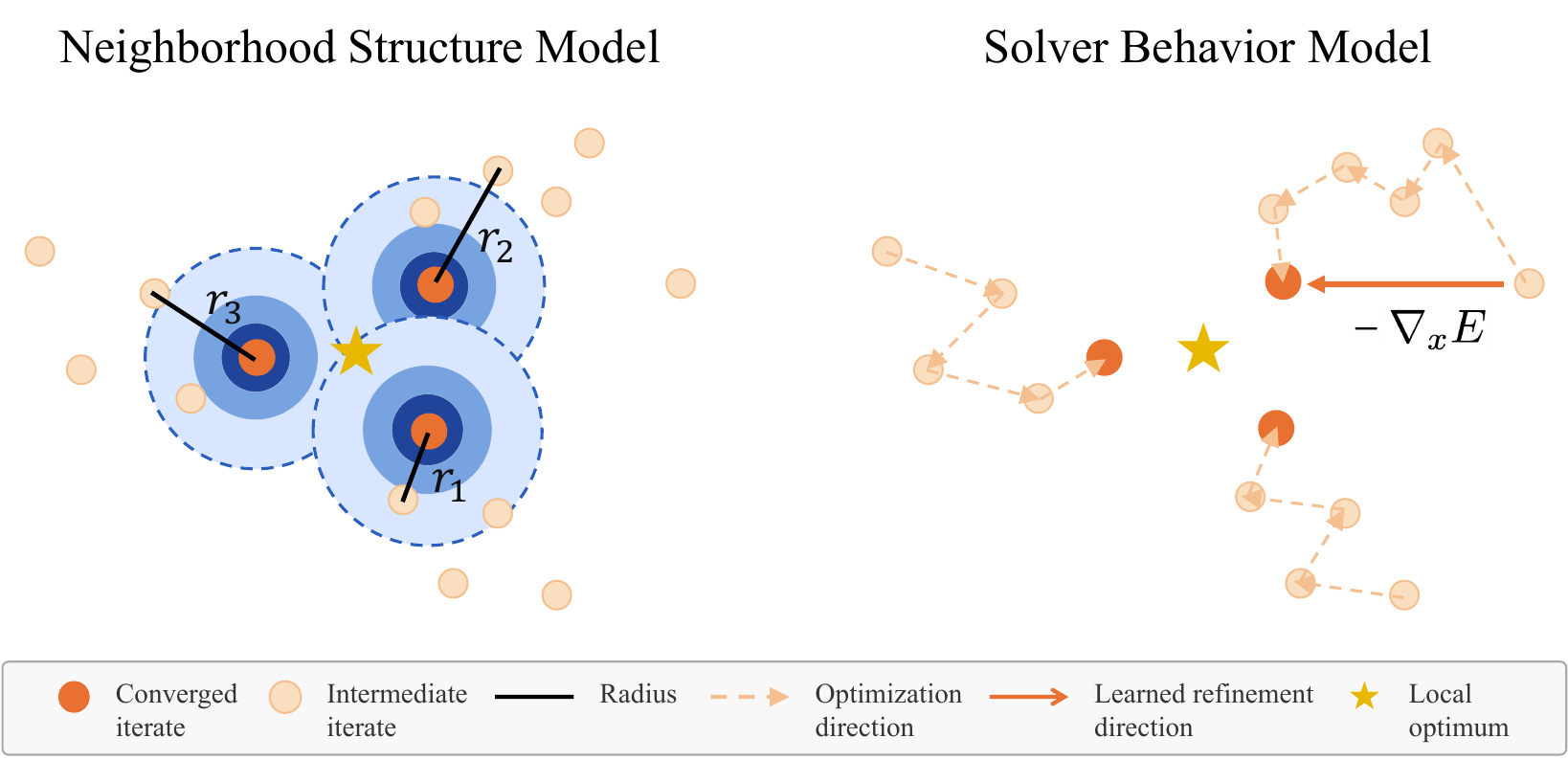}
    \caption{An illustration of the \neighbormodel (left) and \solvermodel (right) in \methodname.
    The \neighbormodel uses a diffusion model to learn the local geometry of the locally optimal solutions, conditioned on the scalar radius $r$ that measures the distance from each intermediate iterate to its converged iterate.
    The \solvermodel learns the refinement direction $- \nabla_x E$ from the solver and guides the \neighbormodel samples toward the optima during the reverse diffusion process.
    }
    \label{fig: methodology}
\end{figure}

In this paper, we propose \methodname (\textbf{G}lobal Search via \textbf{Le}ar\textbf{n}ing from \textbf{S}olver Iterates), a data-driven approach that exploits intermediate iterates to enable robust and data-efficient global search for non-convex continuous optimization.
The method consists of two learning-based components.
The \neighbormodel is a conditional diffusion model trained on the last few iterates prior to convergence, learning the local geometry of optimal solutions and how it varies with problem parameters and relative distance.
The \solvermodel is a neural network that predicts solver refinement directions from intermediate solver iterates, mimicking the solver's refinement behavior with the potential to fast-forward the solver at runtime. 
It provides additional guidance that steers diffusion samples toward high-quality initial guesses; that is, samples that are close to locally optimal solutions and therefore require fewer solver iterations to reach convergence.

We illustrate the proposed framework through a sequence of problems with increasing complexity.
We begin with a soft-constrained quadratic program (QP) solved by gradient descent.
This unimodal example provides a controlled setting in which to examine whether \methodname accurately captures local neighborhood structures and their variation with problem parameters.
This choice is motivated by the observation that, in many non-convex problems, the neighborhood of a local optimum can be well approximated by a constrained quadratic problem.
We then consider several multimodal non-convex benchmark problems, including modified Himmelblau, Rosenbrock, and Levy functions, using the L-BFGS-B solver~\cite{byrd1995limited}.
These examples evaluate the ability of \methodname to learn and sample diverse locally optimal solutions.
We also analyze key hyperparameter choices.
Finally, we apply \methodname to a two-robot navigation problem with nonlinear dynamics in a multi-obstacle environment using SNOPT~\cite{gill2005snopt}.

The paper is organized as follows.
In Section~\ref{sec: background}, we introduce parametric optimization problems and the amortized global search (AmorGS) framework.
We also present the diffusion modeling setup.
In Section~\ref{sec: method}, we introduce the $k$-neighborhood dataset definition and present the two-component learning architecture in \methodname.
It consists of a \neighbormodel based on conditional diffusion models and a \solvermodel with a U-Net architecture.
In Section~\ref{sec: numerical experiment}, we evaluate the proposed method on several numerical examples: a soft-constrained QP and several non-convex benchmark functions, including the modified Himmelblau, Rosenbrock, and Levy functions.
In Section~\ref{sec: hyperparameter}, we analyze how different hyperparameter choices affect performance.
In Section~\ref{sec: real world examples}, we demonstrate the effectiveness of \methodname on a two-robot navigation problem.
Finally, Section~\ref{sec: conclusion} summarizes the contributions and discusses future directions.

\subsection{Related work}

\textbf{Global search.}
For non-convex optimization problems arising in real-world applications, global search is often required to uncover a diverse set of high-quality local optima. 
Evolutionary algorithms, including genetic algorithms~\cite{holland1992genetic} and Differential Evolution~\cite{storn1997differential}, as well as particle swarm optimization~\cite{kennedy1995particle}, represent a class of population-based global search methods.
These methods have been successfully applied in space trajectory design~\cite{cage1994interplanetary,kim2002optimal,olds2007interplanetary,izzo2007search}, circuit design~\cite{miller1997designing,fakhfakh2010analog}, and job scheduling~\cite{kobayashi1995efficient,pezzella2008genetic}.
Hybrid optimization methods, such as multi-start strategies~\cite{ugray2007scatter}, first perform global sampling over the solution space to generate multiple initial guesses and then apply gradient-based numerical solvers to converge to nearby local optima~\cite{locatelli2016global}.
Building on this idea, MBH~\cite{wales1997global,leary2000global} adds random perturbations to local optimization based on the observation that nearby local optima are often connected through funnel-like energy landscapes. 
MBH has been successfully applied in chemical structure optimization~\cite{rondina2013revised,banerjee2021crystal}, bioinformatics~\cite{kucharik2014basin,prentiss2008protein}, and space trajectory design~\cite{englander2017automated,mccarty2018parallel}, outperforming evolutionary algorithms and particle swarm optimization in certain benchmark settings~\cite{vasile2010analysis}.

Despite their effectiveness, these classical global search methods are often computationally expensive, as identifying suitable initial guesses in high-dimensional and highly nonlinear spaces remains challenging. 
Moreover, such methods typically do not exploit data from previously solved problem instances to learn or reuse structural information about the solution landscape, and therefore must perform global search from scratch when presented with new problems.
This limitation motivates learning-based approaches that exploit solver-generated data to amortize the cost of global search across problem instances.

\textbf{Machine learning for optimization.}
Amortized optimization~\cite{amos2023tutorial} learns solution structure from a family of parametric problems offline to accelerate solving new instances online.
For convex problems, this paradigm has proven effective for predicting warm-starts for QPs~\cite{JMLR:v25:23-1174} and model predictive control problems with linear dynamics~\cite{zhang2019safe,chen2022large} using MLPs.
For non-convex optimization, traditional discriminative models, such as decision trees or MLP-based regressors, have been used to learn solution strategies, for example, by predicting active constraint sets or other structural properties that transform the original problem into a simpler formulation~\cite{bertsimas2021voice,cauligi2021coco,bertsimas2025global}. 
However, learning a single deterministic mapping from problem parameters to solutions is often insufficient for highly nonlinear problems with multiple local optima; thus, recent work has explored generative models for learning conditional distributions over locally optimal solutions. 
Graph-based diffusion models have been proposed for combinatorial optimization problems~\cite{sun2023difusco,li2023t2t}. 
In the context of non-convex trajectory optimization, support vector machines~\cite{cassioli2012machine}, variational autoencoders (VAEs)~\cite{li2023amortized}, transformers~\cite{sharony2024learning}, and diffusion models~\cite{li2025diffusolve,li2025aligning,graebner2024learning,graebner2025global} have been used to generate diverse and high-quality initial guesses which can then be refined by numerical solvers to converge to nearby optima.

Different from prior work~\cite{cassioli2012machine,bertsimas2021voice,cauligi2021coco,bertsimas2025global}, which focuses on predicting the globally optimal solution or improving the quality of a single solution over existing solvers, generative model-based approaches~\cite{sharony2024learning,li2025diffusolve,graebner2025global} aim to learn distributions over multiple locally optimal solutions and how they vary with problem parameters.
This perspective is particularly appealing in applications where a diverse set of high-quality solutions is desired for downstream decision-making.

Despite their promise, generative model-based approaches to non-convex optimization face a significant challenge: data scarcity.
Constructing training datasets typically requires expensive global search using numerical solvers such as SNOPT~\cite{gill2005snopt} or iLQR~\cite{li2004iterative}, run repeatedly across many problem instances~\cite{li2025diffusolve,sharony2024learning}.
Moreover, while solver-generated data is abundant during the optimization process, most existing methods retain only the final converged solutions and discard intermediate solver iterates.
Although recent work has begun to connect the optimization process with diffusion training dynamics~\cite{giannone2023aligning}, learning the local neighborhood structure around optima and explicitly exploiting solver behaviors remain largely unexplored.

\textbf{Diffusion models.}
Diffusion models are a class of generative models inspired by non-equilibrium thermodynamics, in which random noise is gradually added to training data through a forward diffusion process, and a reverse denoising process is learned to recover data from noise~\cite{sohl2015deep,song2019generative,ho2020denoising}. 
With their strong capability in modeling complex, high-dimensional distributions, diffusion models have been successfully applied to a wide range of domains, including image and video generation~\cite{nichol2021glide,rombach2022high,ho2022video}, protein structure generation~\cite{jing2023eigenfold,wu2024protein,schneuing2024structure}, and behavior modeling for robotics and autonomous vehicles~\cite{janner2022planning,ajay2022conditional,chi2025diffusion,jiang2023motiondiffuser,zhang2024predicting,li2025predictive}.
To enable conditional generation, classifier-free guidance is commonly used in diffusion models by jointly training conditional and unconditional models and interpolating their predictions during sampling~\cite{ho2022classifier}.
In contrast, classifier guidance trains a separate classifier or energy model on noisy samples and injects its gradient into the diffusion sampling process to steer generation toward desired conditions~\cite{dhariwal2021diffusion}.
Regarding model architectures, U-Nets~\cite{ronneberger2015u} are commonly used for modeling sequential data~\cite{janner2022planning}, and transformers~\cite{vaswani2017attention} are popular in image and video domains~\cite{peebles2023scalable}.

\subsection{Contributions}

Our major contributions are as follows:

\begin{itemize}
    \item \textbf{Data augmentation from solver iterates.}
    During each solver run, instead of collecting only the final converged solutions, we retain the last few solver iterates before convergence.
    This provides free data augmentation that enriches the dataset with local geometry information around optima, without additional computational cost.
    \item \textbf{GLENS method.}
    We propose \methodname, a diffusion model-based global search method that exploits the augmented dataset through two complementary components: a \neighbormodel that learns the distribution of local geometry around optima conditioned on problem parameters and relative distances, and a \solvermodel that captures solver refinement directions to further guide diffusion samples toward nearby optima.
    \item \textbf{Experiments and analysis.}
    Through numerical experiments of increasing complexity, including a soft-constrained QP, several non-convex benchmark functions, and a two-robot navigation problem, we show that \methodname generates high-quality initial guesses while preserving solution diversity, leading to improved solver convergence across different problem settings and solvers.
    We also analyze key hyperparameter choices from both data and modeling perspectives, which provides practical guidance when applying to new problem families.
\end{itemize}

\section{Background}\label{sec: background}

\subsection{Parametric optimization}

We consider a family of parametric continuous optimization problems $\Omega$, where each problem instance $\mathcal{P}_{\alpha} \in \Omega$ is specified by a parameter $\alpha \in \mathbb{R}^{d_{\alpha}}$ and takes the form
\begin{equation} \label{eq: parameterized optimization}
    \begin{array}{ll}
        \underset{x \in \mathbb{R}^{d_x}}{\text{minimize}}   & f(x; \alpha) \\
        \text{subject to} & g(x; \alpha) \leq 0, \\
                          & h(x; \alpha) = 0.
    \end{array}
    \tag{$\mathcal{P}_\alpha$}
\end{equation}
Here, $x \in \mathbb{R}^{d_x}$ is the continuous decision variable and $\alpha \in \mathbb{R}^{d_{\alpha}}$ is the problem parameter.
The objective $f \colon \mathbb{R}^{d_x} \times \mathbb{R}^{d_\alpha} \to \mathbb{R}$, the inequality constraint function $g \colon \mathbb{R}^{d_x} \times \mathbb{R}^{d_\alpha} \to \mathbb{R}^{m_g}$, and the equality constraint function $h \colon \mathbb{R}^{d_x} \times \mathbb{R}^{d_\alpha} \to \mathbb{R}^{m_h}$ have fixed functional forms across all instances, with the inequality $g(x; \alpha) \leq 0$ understood componentwise; the parameter $\alpha$ specifies the problem data of each instance.
We assume that, for every $\alpha \in \mathbb{R}^{d_\alpha}$, the functions $f$, $g$, and $h$ are smooth in $x$, but allow them to be non-convex.

The first-order Karush-Kuhn-Tucker (KKT) optimality conditions~\cite{Kuhn2014} for problem~\eqref{eq: parameterized optimization} can be written directly in residual form, with Lagrange multipliers $\lambda \in \mathbb{R}^{m_g}$ and $\nu \in \mathbb{R}^{m_h}$ for the inequality and equality constraints.
For a tolerance $\varepsilon_{\text{KKT}} > 0$, we say that $x$ satisfies the KKT conditions if there exist multipliers $\lambda$ and $\nu$ such that
\begin{equation}\label{eq: kkt}
\begin{aligned}
\|\nabla_x f(x; \alpha) + \nabla_x g(x; \alpha)^\top \lambda + \nabla_x h(x; \alpha)^\top \nu\|_\infty &\leq \varepsilon_{\text{KKT}}, \\
\|h(x; \alpha)\|_\infty &\leq \varepsilon_{\text{KKT}}, \\
\|\max\{g(x; \alpha),\, 0\}\|_\infty &\leq \varepsilon_{\text{KKT}}, \\
\|\max\{-\lambda,\, 0\}\|_\infty &\leq \varepsilon_{\text{KKT}}, \\
|\lambda^\top g(x; \alpha)| &\leq \varepsilon_{\text{KKT}},
\end{aligned}
\end{equation}
where $\nabla_x g(x; \alpha)$ and $\nabla_x h(x; \alpha)$ are the Jacobians of $g$ and $h$ with respect to $x$, and $\max\{\cdot,\, 0\}$ is taken componentwise.
The five inequalities correspond, respectively, to stationarity, primal equality feasibility, primal inequality feasibility, dual feasibility, and complementarity.

To solve a non-convex problem instance $\mathcal{P}_{\alpha}$, classical global search
methods typically rely on two components: (i) an initial guess generator $\Gamma$ and
(ii) a gradient-based numerical solver $\pi$~\cite{locatelli2016global}.
The generator $\Gamma$ defines a global sampling distribution over the decision space $\mathbb{R}^{d_x}$, possibly conditioned on the problem parameter $\alpha$, from which an initial guess $\mathbb{R}^{d_x} \ni x^{0} \sim \Gamma$ is drawn.
We model the numerical solver as a single-step update operator $\pi: \mathbb{R}^{d_x} \times \mathbb{R}^{d_\alpha} \to \mathbb{R}^{d_x}$ that, starting from $x^{0}$, maps each iterate to the next:
\begin{align}\label{eq: solver_update}
    x^{m+1} = \pi(x^{m}; \alpha), \quad m = 0, 1, \ldots,
\end{align}
until a termination criterion is satisfied.
If the solver converges after $n$ iterations and the final iterate $x^{n}$ satisfies the KKT conditions~\eqref{eq: kkt} to tolerance $\varepsilon_{\text{KKT}}$, we call $x^{n}$ a \emph{locally optimal solution} and denote it $x^*(x^{0}; \pi, \alpha)$.
This notation emphasizes the dependence on the initial guess $x^{0}$, the solver $\pi$, and the problem parameter $\alpha$.

Since different initial guesses $x^{0}$ may converge to different local optima, obtaining a diverse collection of locally optimal solutions typically requires sampling many initial guesses from $\Gamma$ and running the local search with $\pi$ from each initial guess.
This global search process can be computationally expensive for high-dimensional and highly nonlinear problems.

\paragraph{Notation.}
Throughout the paper, superscripts on $x$ denote solver iteration indices, not exponents: 
$x^0$ is the initial guess, $x^m$ is the iterate after $m$ solver steps, and $x^*$ is the locally optimal solution should the initial guess converge. 
For brevity, we will often suppress the dependence of $x^*$ on the initial condition $x^0$, solver $\pi$, and parameter $\alpha$.

\subsection{Amortized global search}\label{sec: amorGS}

AmorGS~\cite{li2023amortized,beeson2024global} is a data-driven approach that aims to reduce the computational cost of sampling diverse and high-quality initial guesses for parametric optimization problems.
Here, high-quality initial guesses refer to samples that are close to optima and therefore require fewer solver iterations to converge.
Rather than treating each problem instance $\mathcal{P}_{\alpha} \in \Omega$ independently, AmorGS uses data from a family of solved instances in $\Omega$ to learn how locally optimal solutions vary with respect to the problem parameter $\alpha$.

For offline data collection, AmorGS aims to construct a local optimum dataset $\mathcal{D}^*$, which consists of pairs of parameters and locally optimal solutions as follows.

\begin{definition}[Local optimum dataset]\label{def: local optimum dataset}
Let $\{\mathcal{P}_{\alpha_i}\}_{i=1}^N$ be $N$ problem instances sampled from the parametric problem family $\Omega$ in problem~\eqref{eq: parameterized optimization}.
For each parameter $\alpha_i$, draw $M$ initial guesses $\{x^{0}_{i,j}\}_{j=1}^M$ from an initial guess generator $\Gamma$ and run the solver iteration in Eq.~\eqref{eq: solver_update} from each one until numerical termination (convergence or otherwise).
Let $\mathcal{O}_i \subseteq \{1, \ldots, M\}$ be the set of indices for which the solver converges, and write
\begin{align}\label{eq: x star indexed}
    x^*_{i,j} \coloneqq x^*(x^{0}_{i,j}; \pi, \alpha_i)
\end{align}
for the resulting locally optimal solution.
The \emph{local optimum dataset} is the collection of parameter--solution pairs
\begin{align}\label{eq: local optimum dataset}
    \mathcal{D}^* \coloneqq \bigl\{ (\alpha_i,\, x^*_{i,j}) \mid i = 1, \ldots, N,\; \forall j \in \mathcal{O}_i \bigr\}.
\end{align}
\end{definition}

A generative model is trained on $\mathcal{D}^*$ to sample from the conditional distribution of locally optimal solutions $x^*_{i,j}$ given the corresponding problem parameter $\alpha_i$.
During testing, the trained generative model serves as a learned initial guess generator $\hat{\Gamma}_{\alpha}$ conditioned on a new parameter value $\alpha$, producing high-quality and diverse initial guesses. 
Because these generated initial guesses are close to locally optimal solutions, the numerical solver $\pi$ can converge quickly, significantly reducing the cost of global search for previously unseen problem instances.
However, when the solving process is computationally expensive, this standard AmorGS may face the training data scarcity issue and also neglect valuable neighborhood information around the optima.

Prior work has shown that diffusion models are particularly effective in the AmorGS framework, outperforming alternative generative approaches such as VAEs~\cite{li2025diffusolve}.
We detail the main features of diffusion models in the next section before introducing their use in the modeling contributions of this paper in Section~\ref{sec: method}.

\subsection{Diffusion models}\label{sec: diffusion models}

Diffusion models consist of a forward process to generate noisy data for training and a learnable reverse process for data sampling at test time.
Within the AmorGS framework~\cite{li2025diffusolve}, denoising diffusion probabilistic models (DDPM)~\cite{ho2020denoising} are used as the data-driven initial guess generator $\hat{\Gamma}_{\alpha}$ conditioned on parameter $\alpha$.

\textbf{Training.}
Given a data sample $z \sim q_{\text{data}}$ drawn from the data distribution, e.g.,\ the distribution of locally optimal solutions, the forward diffusion process gradually adds Gaussian noise to the data.
Let $z_0 \coloneqq z$; the forward process generates a sequence of noisy samples $z_1, z_2, \ldots, z_T$ with $q(z_t|z_{t-1})=\mathcal{N}(z_t ; \sqrt{1-\beta_t} z_{t-1}, \beta_t I)$, where $0 < \beta_1 < \beta_2 < \cdots < \beta_T = 1$ is an increasing noise schedule.
With the reparameterization trick~\cite{ho2020denoising} and letting $\sigma_t \coloneqq 1 - \beta_t$ and $\bar\sigma_t \coloneqq \prod_{i=1}^{t} \sigma_i$, the noisy sample $z_t$ has the closed-form expression
\begin{align}\label{eq: diffusion forward sample closed form}
    z_t = \sqrt{\bar\sigma_t}\, z_0 + \sqrt{1 - \bar\sigma_t}\, \epsilon,
\end{align}
where $\epsilon \sim \mathcal{N}(0, I)$.

In the reverse process, DDPM starts from a sample $z_T \sim \mathcal{N}(0, I)$ and learns a model $p_{\theta}$, parameterized by $\theta$, that iteratively denoises the data as $p_{\theta}(z_{t-1} \mid z_t) = \mathcal{N}(z_{t-1}; \mu_{\theta}(z_t, t), \beta_t I)$.
Instead of predicting $\mu_{\theta}(z_t, t)$ directly, DDPM predicts the noise $\epsilon_{\theta}(z_t, t)$ added to $z_t$, with the transformation
\begin{align}\label{eq: diffusion predicts mu by noise}
    \mu_{\theta}(z_t, t) = \frac{1}{\sqrt{\sigma_t}} \left( z_t - \frac{1 - \sigma_t}{\sqrt{1 - \bar\sigma_t}}\, \epsilon_{\theta}(z_t, t) \right).
\end{align}
Since $z_t$ can be expressed in closed form using Eq.~\eqref{eq: diffusion forward sample closed form}, the DDPM training loss is
\begin{align}\label{eq: diffusion loss functions}
    \mathcal{L}_{\text{DDPM}} = \mathbb{E}_{z_0, \epsilon, t} \left\| \epsilon - \epsilon_{\theta}\!\left( \sqrt{\bar\sigma_t}\, z_0 + \sqrt{1 - \bar\sigma_t}\, \epsilon,\, t \right) \right\|^2,
\end{align}
where $z_0 \sim q_{\text{data}}$, $\epsilon \sim \mathcal{N}(0, I)$, and $t$ is uniformly drawn from $[1, T]$.

\textbf{Sampling.}
Sampling begins by drawing $z_T \sim \mathcal{N}(0, I)$, followed by iterative denoising steps 
\begin{align}\label{eq: ddpm sampling}
    z_{t-1} \sim \mathcal{N}(\mu_{\theta}(z_t, t), \beta_t I), \quad t = T-1, T-2, \dots, 1,
\end{align}
where $\mu_{\theta}(z_t, t)$ is derived from the learned noise predictor $\epsilon_{\theta}(z_t, t)$ using Eq.~\eqref{eq: diffusion predicts mu by noise}.
This procedure is repeated until the clean sample $z_0$ is obtained.

\textbf{Conditional generation.}
Diffusion models support conditional data generation given auxiliary information $y$ associated with the data $z$.
Classifier-free guidance~\cite{ho2022classifier} jointly learns a conditional noise predictor $\epsilon_{\theta}(z_t, t, y)$ and an unconditional predictor $\epsilon_{\theta}(z_t, t, y = \varnothing)$ using the same loss function in Eq.~\eqref{eq: diffusion loss functions} by randomly masking the condition $y$ during training with a certain probability.
During sampling, to generate data $z$ with auxiliary $y$, an interpolated noise predictor $\bar \epsilon$ is used
\begin{align}\label{eq: classifier free noise predictor}
    \bar \epsilon_{\theta}(z_t, t, y) = (s+1) \epsilon_{\theta}(z_t, t, y) - s \epsilon_{\theta}(z_t, t, y=\varnothing),
\end{align}
where $s$ controls the weight of guidance and balances between sample diversity and fidelity.

Classifier guidance~\cite{dhariwal2021diffusion} trains an auxiliary classifier on noisy samples $z_t$ and uses the gradient of its log-likelihood to steer the reverse diffusion process.
More generally, this guidance can be written in the form of an energy function $E_{\phi}(z_t,\alpha)$.
During sampling, the gradient of the energy $\nabla_{z_t} E_{\phi}(z_t, \alpha)$ is injected into each denoising step to steer the diffusion process toward lower energy:
\begin{align}\label{eq: classifier sampling}
    z_{t-1} \sim \mathcal{N}\bigl( \mu_{\theta} - \tilde s\, \beta_t \nabla_{z_t} E_{\phi}(z_t, \alpha),\, \beta_t I \bigr),
\end{align}
where $\tilde s$ controls the strength of the guidance.

\textbf{Neural network backbone.}
Within the AmorGS framework, the conditional noise predictor $\epsilon_{\theta}(z_t, t, y)$ is implemented using a U-Net architecture~\cite{ronneberger2015u}, following the implementation in~\cite{lucidrains2025denoisingdiffusionpytorch}.
We use the same U-Net architecture as the backbone for the learned gradient field in classifier guidance in this manuscript. 
This architecture is introduced in Section~\ref{sec: solver behavior model}.

\section{Method}\label{sec: method}

We first introduce the concept of a $k$-neighborhood path, inspired by Beeson et al.~\cite{beeson2024global}, which consists of solver iterates in the
neighborhood of a converged local optimum within the same solver run, and define the corresponding $k$-neighborhood dataset.
To exploit the $k$-neighborhood dataset, we then present \methodname, a novel data-driven global search based on diffusion models.
Through a \neighbormodel and a \solvermodel, \methodname enables data-efficient
generation of high-quality initial guesses for non-convex continuous
optimization problems.

\subsection{$k$-neighborhood dataset}

Existing amortized optimization methods typically train on the local optimum dataset $\mathcal{D}^*$ as defined in Definition~\ref{def: local optimum dataset}, which contains only the final converged iterates produced by numerical solvers.
However, our key insight is that the last few solver iterates leading to convergence encode rich information about the neighborhood structure of a locally optimal solution, which is highly informative for data-driven global search.

\begin{figure}[t]
    \centering
    \includegraphics[width=0.4\textwidth]{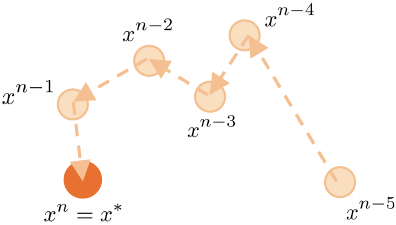}
    \caption{
    Illustration of a local optimum $x^*$ and the corresponding $6$-neighborhood path $\mathcal{N}^6=\{x^{n-5}, x^{n-4}, x^{n-3}, x^{n-2}, x^{n-1}, x^n \}$ within the same solver run.
    }
    \label{fig: k neighborhood path}
\end{figure}

Motivated by this, we introduce the concept of a $k$-neighborhood path, which denotes the final $k$ solver iterates before convergence.
An illustration of a $k$-neighborhood path is shown in Figure~\ref{fig: k neighborhood path}. 

\begin{definition}[$k$-neighborhood path]
\label{def: k neighborhood}
Let $\mathcal{P}_\alpha$ be a problem instance, $x^0$ an initial guess, and $(x^{0}, x^{1}, \ldots, x^{n})$ the sequence of solver iterates converging to a locally optimal solution $x^n = x^*$. For $k \in \mathbb{Z}_{>0}$ with $k \le n + 1$, the \emph{$k$-neighborhood path} is defined as the last $k$ iterates of this solver run:
\begin{align}
    \mathcal{N}^k \coloneqq \bigl\{ x^{n-k+1}, x^{n-k+2}, \ldots, x^{n} \bigr\}. 
\end{align}
We use additional subscripts on $\mathcal{N}^k$ to indicate a path associated with the $i$-th problem instance $\mathcal{P}_{\alpha_i}$ and a solver initialized with the $j$-th initial guess $x^0_{i,j}$; in particular, this would yield the definition:
\begin{align}
    \mathcal{N}^k_{i,j} \coloneqq
    \bigl\{ x^{n-k+1}_{i,j}, x^{n-k+2}_{i,j}, \ldots, x^{n}_{i,j} \bigr\}.
\end{align}
\end{definition}

The $1$-neighborhood path $\mathcal{N}^1 = \{x^*\}$ reduces to the locally optimal solution itself.
In Figure~\ref{fig: k neighborhood path}, we show an example of a local optimum $x^*$ and its corresponding $6$-neighborhood path $\mathcal{N}^6$.

Building on the $k$-neighborhood path concept, we introduce the following $k$-neighborhood dataset $\mathcal{D}^k$.
\begin{definition}[$k$-neighborhood dataset]\label{def: k neighborhood dataset}
Under the setup of Definition~\ref{def: local optimum dataset}, let $\mathcal{N}^k_{i,j}$ denote the $k$-neighborhood path of the solver run
initialized at $x^{0}_{i,j}$, as defined in Definition~\ref{def: k neighborhood}.
The \emph{$k$-neighborhood dataset} is the collection of triples
\begin{align}
    \mathcal{D}^k \coloneqq \bigl\{
        (\alpha_i,\, x,\, x^*_{i,j})
        \mid
        i = 1, \ldots, N,\;
        \forall j \in \mathcal{O}_i,\;
        \forall x \in \mathcal{N}^k_{i,j}
    \bigr\}.
\end{align}
Each data point consists of a problem parameter, an iterate from the corresponding $k$-neighborhood path, and the locally optimal solution reached by that solver
run.
\end{definition}

By construction, the $1$-neighborhood dataset $\mathcal{D}^1$ corresponds to the local optimum dataset $\mathcal{D}^*$ in Definition~\ref{def: local optimum dataset}.
Furthermore, the $k$-neighborhood dataset $\mathcal{D}^k$ contains up to $k$ solver iterates associated with each locally optimal solution, without requiring additional solver runs.
These additional iterates provide richer information about the local neighborhood structure surrounding each locally optimal solution.

\subsection{Two-component learning architecture}

Although the $k$-neighborhood dataset $\mathcal{D}^k$ substantially enriches the available training data, the intermediate solver iterates are generally suboptimal, making them challenging for direct use in existing data-driven optimization methods. 
To better exploit the $k$-neighborhood dataset, we propose \methodname, a global search method using diffusion models that consists of two complementary learning components: a \neighbormodel that captures the neighborhood geometry and a \solvermodel that learns the solver refinement behavior.
The two models are trained separately offline on $\mathcal{D}^k$ and combined during online sampling to guide diffusion-based initial guess generation.

\subsubsection{Neighborhood structure model}

The \neighbormodel is a conditional diffusion model that serves as a data-driven
initial guess generator $\hat{\Gamma}_{\alpha}$.
Different from the generator in Section~\ref{sec: amorGS}, it is trained on the $k$-neighborhood dataset $\mathcal{D}^k$ from Definition~\ref{def: k neighborhood dataset}, whose entries are triples $(\alpha, x, x^*)$ with $x$ being a solver iterate in the $k$-neighborhood path of an optimum $x^*$.
This enlarges the support of the empirical training data distribution beyond locally
optimal solutions.
However, when conditioned only on the problem parameter $\alpha$, all iterates within a neighborhood are treated equivalently, preventing the model from capturing their relative geometric structure.

To encode this relative structure, we augment the conditioning information with an
additional scalar variable $r$, the distance between a solver iterate
and the locally optimal solution reached by the same solver run.
For each data point $(\alpha, x, x^*)$ in the $k$-neighborhood dataset $\mathcal{D}^k$, we introduce
\begin{align}\label{eq: distance to optimum}
    r(x,x^*) \coloneqq \bigl\| x - x^* \bigr\|_2 .
\end{align}
The diffusion-based \neighbormodel is then trained to sample solver iterates
conditioned on the augmented variable $y \;\coloneqq\; (\alpha, r)$, where $\alpha$ captures problem-level information and $r$ encodes the relative position of
an iterate within the local neighborhood.

At test time, when a new problem instance $\mathcal{P}_{\alpha}$ is given,
its locally optimal solutions are unknown.
Querying the \neighbormodel with $r = 0$ biases the diffusion sampling process toward
samples concentrated near the manifold of locally optimal solutions.
In this way, the \neighbormodel naturally serves as a learned initial guess generator
$\hat{\Gamma}_{\alpha}$ conditioned on parameter $\alpha$ for data-driven global search.

Training and sampling follow the standard DDPM framework reviewed in Section~\ref{sec: diffusion models}, with classifier-free guidance for the conditional variable $y$.
At training time, an unconditional noise predictor $\epsilon_\theta(z_t, t, y = \varnothing)$ and a conditional noise predictor $\epsilon_\theta(z_t, t, y)$, parameterized by a neural network with weights $\theta$, are jointly trained on the $k$-neighborhood dataset $\mathcal{D}^k$ using the DDPM objective in Eq.~\eqref{eq: diffusion loss functions}; the conditioning variable $y$ is randomly dropped with probability $p_{\text{uncond}}$.
At sampling time, the guided noise predictor $\bar\epsilon_\theta(z_t, t, y)$ in Eq.~\eqref{eq: classifier free noise predictor} with weight $s_{\text{NS}}$ is plugged into Eqs.~\eqref{eq: diffusion predicts mu by noise} and~\eqref{eq: ddpm sampling} to iteratively denoise from Gaussian noise.

\subsubsection{Solver behavior model}\label{sec: solver behavior model}

While the \neighbormodel captures the geometry of solution neighborhoods from a global perspective, it does not explicitly model how the numerical solver locally refines iterates toward nearby optima.
The generated samples may therefore still deviate from exact optima due to generalization error, even when queried with $r=0$.

To address this limitation, we introduce the \solvermodel, which learns the local refinement behavior of the numerical solver from $k$-neighborhood paths and is used to further guide diffusion samples toward their associated optima.
For each data point $(\alpha,x,x^*)$ in the $k$-neighborhood dataset $\mathcal{D}^k$, we define the energy function
\begin{align}
    E(x,x^*)
    \coloneqq
    \frac{1}{2} r(x, x^*)^2
    =
    \frac{1}{2}
    \bigl\|x-x^*\bigr\|_2^2 .
\end{align}
Its gradient with respect to $x$ is
\begin{align}\label{eq: gradient of energy}
    \nabla_x E(x,x^*) = x - x^* .
\end{align}
Thus, the negative gradient $-\nabla_x E(x,x^*)$ points from the iterate $x$ toward the associated locally optimal solution.
The gradient of this energy function characterizes the solver refinement behavior of each iterate within the $k$-neighborhood path.

The \solvermodel is parameterized as a neural network
$\xi_{\phi}: \mathbb{R}^{d_x} \times \mathbb{R}^{d_{\alpha}} \to \mathbb{R}^{d_x}$ and is trained to predict this gradient field.
During training, for each entry $(\alpha,x,x^*) \in \mathcal{D}^k$, the optimum $x^*$ is used only to construct the supervised target.
Specifically, the network $\xi_\phi$ is learned by minimizing
\begin{align}
\mathcal{L}_{\text{SB}}
\;=\;
\mathbb{E}_{(\alpha,x,x^*)\sim\mathcal D^k}
\Bigl[
\bigl\|
\xi_{\phi}(x,\alpha)
-
\nabla_x E(x,x^*)
\bigr\|_2^2
\Bigr].
\end{align}
At test time, $\xi_\phi$ predicts the gradient field from $(x,\alpha)$ alone.

Standard classifier guidance trains an energy function on Gaussian-corrupted samples and obtains the guidance direction by differentiating its output.
The \solvermodel instead directly learns a gradient field from structured $k$-neighborhood path iterates around local optima.
These iterates can be viewed as solver-generated perturbations around their corresponding local optima.
During online sampling, we set the radius $r=0$ in the \neighbormodel to generate samples near local optima, and apply the \solvermodel guidance only in the final denoising steps $t \le t_{\text{guide}}$.
At these late steps, the diffusion latent state $z_t$ is only lightly corrupted and is assumed to lie near the learned neighborhood around a local optimum.
Thus, evaluating $\xi_\phi(z_t,\alpha)$ can be approximated by the learned gradient field $\xi_\phi(x,\alpha)$ trained on the $k$-neighborhood dataset.
Following the classifier-guidance form in Eq.~\eqref{eq: classifier sampling}, this refinement gradient field is incorporated with strength $\tilde{s}=s_{\text{SB}}$ to guide the sample toward the corresponding optimum.

\begin{algorithm}[t]
\caption{GLENS online sampling}
\label{alg:glens_sample}
\begin{algorithmic}[1]
\Require \neighbormodel noise predictor $\epsilon_{\theta}$ with weight $s_{\text{NS}}$
\Require \solvermodel gradient field $\xi_{\phi}$ with weight $s_{\text{SB}}$
\Require Diffusion steps $T$, guidance activation step $t_{\text{guide}}$
\Require Noise schedule $\{\beta_t\}_{t=1}^T$, $\sigma_t \coloneqq 1-\beta_t$, $\bar\sigma_t \coloneqq \prod_{i=1}^t \sigma_i$
\Require Test parameter $\alpha$
\Ensure Initial guess prediction $\hat{x}$
\State Set $r \gets 0$ and $y \gets (\alpha,r)$.
\State Sample $z_T \sim \mathcal{N}(0,I)$.
\For{$t = T, T-1, \ldots, 1$}
    \State Compute classifier-free guided noise
    \begin{align}
        \bar\epsilon_\theta(z_t,t,y)
        \gets
        (1+s_{\text{NS}})\epsilon_\theta(z_t,t,y)
        -
        s_{\text{NS}}\epsilon_\theta(z_t,t,\varnothing). \nonumber
    \end{align}
    \State Compute mean $\mu_\theta(z_t,t,y)$
    \begin{align}
        \mu_{\theta}(z_t,t,y)
        \gets
        \frac{1}{\sqrt{\sigma_t}}
        \left(
            z_t
            -
            \frac{1-\sigma_t}{\sqrt{1-\bar\sigma_t}}\,
            \bar\epsilon_{\theta}(z_t,t,y)
        \right). \nonumber
    \end{align}
    \If{$t \le t_{\text{guide}}$}
        \State Approximate the solver refinement direction by $-\xi_\phi(z_t,\alpha)$.
        \State Sample with solver refinement
        \begin{align}
            z_{t-1}
            \sim
            \mathcal{N}\!\left(
                \mu_\theta(z_t,t,y)
                -
                s_{\text{SB}}\,\beta_t\,\xi_\phi(z_t,\alpha),\;
                \beta_t I
            \right). \nonumber
        \end{align}
    \Else
        \State Sample without solver refinement
        \begin{align}
            z_{t-1}
            \sim
            \mathcal{N}\!\left(
                \mu_\theta(z_t,t,y),\;
                \beta_t I
            \right). \nonumber
        \end{align}
    \EndIf
\EndFor
\State \Return $\hat{x} \gets z_0$
\end{algorithmic}
\end{algorithm}

\subsubsection{Combined online sampling}
While the \neighbormodel and \solvermodel are trained separately offline, they are coupled during online testing to perform a guided
diffusion sampling process.
Combining neighborhood structure modeling with
solver refinement enables high-quality initial guess generation for data-driven global search in \methodname.
For clarity, we summarize the online sampling procedure of
\methodname in Algorithm~\ref{alg:glens_sample}.

\section{Numerical experiments}\label{sec: numerical experiment}

We evaluate \methodname through numerical experiments designed to investigate how learning from $k$-neighborhood dataset with the \neighbormodel and \solvermodel improves data-driven global search.
In particular, the experiments assess
(i) whether \methodname enables more accurate and robust modeling of how locally optimal solutions vary with problem parameters, and
(ii) whether this capability generalizes to non-convex problems with multiple local optima.
Throughout this section, when a problem instance has a multi-component parameter, we write $\alpha = (\alpha_1, \alpha_2, \ldots) \in \mathbb{R}^{d_\alpha}$, with subscripts denoting components of a single parameter vector rather than indices over problem instances.

\textbf{Test problems.}
We first consider a baseline soft-constrained QP solved with gradient descent.
This unimodal example mimics the neighborhood structure around a single optimum.
It establishes how all compared methods learn the solution variation with respect to problem parameters.
We then extend our evaluation to multimodal non-convex benchmarks, including the modified Himmelblau function, Rosenbrock function, and Levy function, solved with the L-BFGS-B solver, demonstrating how \methodname enables effective global search in the presence of multiple distinct optima.

\textbf{Model setup.}
We use ``NS'' to denote the \neighbormodel and ``SB'' to denote the \solvermodel.
The \neighbormodel uses a diffusion model with $50$ sampling steps and a U-Net backbone for the noise predictor.
The U-Net uses a base feature dimension of $64$ and three downsampling and upsampling layers with channel multipliers of $\{1,2,4\}$.
The \solvermodel also uses a U-Net backbone to learn the gradient field, with similar three layers for downsampling and upsampling, with base feature dimension of $64$ and channel multipliers of $\{1,2,4\}$.

\begin{figure}[t]
    \centering
    \includegraphics[width=0.85\textwidth]{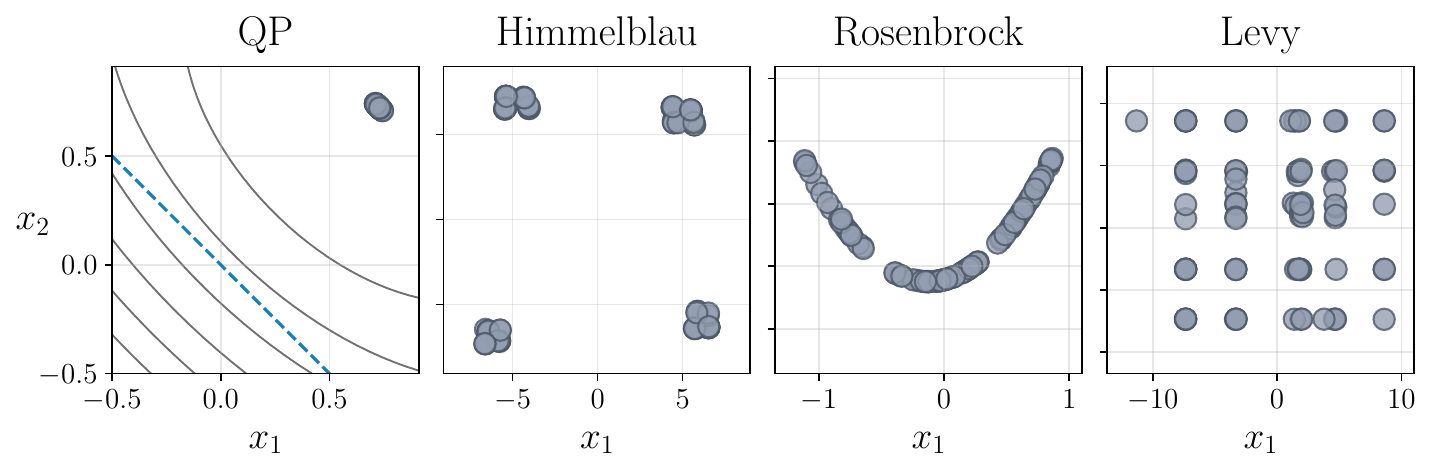}
    \caption{
    Ground truth solutions for the test problems in the first two dimensions, shown for a fixed parameter value. 
    Each grey circle represents a converged local optimum obtained by the corresponding solver.
    For the QP example, objective contours are shown with the constraint indicated by the blue dashed line.
    The non-convex problems, including Himmelblau, Rosenbrock, and Levy, exhibit rich multimodal structures with multiple local optima.
    }
    \label{fig: groundtruth sample visualization}
\end{figure}

During testing, we set the guidance weight $s_{\text{NS}} = 0.5$ for \neighbormodel in classifier-free guidance, and apply the solver refinement direction from \solvermodel with weight $s_{\text{SB}} = 100$ when diffusion step $t \leq t_{\text{guide}} = 5$.

\textbf{Comparison methods.}
We compare the following methods that generate initial guesses for global search:
\begin{itemize}
    \item \textbf{Uniform}: Initial guesses are sampled uniformly from the solution space.
    \item \textbf{DiffuSolve}~\cite{li2025diffusolve}: A state-of-the-art diffusion-based global search method trained only on $1$-neighborhood data, i.e., the locally optimal solutions.
    To ensure a fair comparison, it uses the same U-Net architecture as our \neighbormodel.
    \item \textbf{DiffuSolve-k}: The same model used in DiffuSolve~\cite{li2025diffusolve} but trained on the $k$-neighborhood dataset.
    \item \textbf{NS-only} (\methodname without solver guidance): The proposed \neighbormodel trained on the $k$-neighborhood dataset, without the \solvermodel.
    \item \textbf{\methodname}: The full proposed method that combines the \neighbormodel and \solvermodel, both trained on the $k$-neighborhood dataset.
\end{itemize}

\textbf{Evaluation metrics.}
We evaluate the quality of sampled initial guesses for global search mainly from the following two aspects.
\begin{itemize}
    \item \textbf{Convergence behavior.}
For an unseen problem instance, we measure which $k$-neighborhood the generated initial guesses lie in, equivalently, the number of iterations required at most by the solver to converge.
Initial guesses that fall into smaller $k$-neighborhoods lead to faster solver convergence and indicate higher quality predictions conditioned on the problem parameters.
    \item \textbf{Diversity and coverage.}
For non-convex problems with multiple local optima, we examine whether the generated samples cover multiple distinct ground truth optima rather than collapsing to a single mode.
\end{itemize}
The sample visualizations and the $k$-neighborhood statistics provide complementary views of performance.
The figures are intended to qualitatively assess whether the generated samples capture the structure and diversity of the ground truth local optima.
The tables provide the primary quantitative comparison of sample quality by measuring how close the generated samples are to solver convergence.
Therefore, in some examples, different methods may capture similar qualitative solution structures while their differences in sample accuracy are more clearly reflected by the $k$-neighborhood statistics.

\subsection{Baseline problem: soft-constrained QP}

\subsubsection{Problem setup}

We consider the following soft-constrained QP:
\begin{align}\label{eq: qp soft constrained objective}
\min_{x \in \mathbb{R}^{d_x}} \quad
\frac{1}{2} x^\top x
+ \alpha_1 \sum_{i=1}^{d_x/2}
\ln\!\left( 1 + \exp\!\left( - x_{2i-1} - x_{2i} \right) \right).
\end{align}
The objective consists of a quadratic term and a soft constraint via the softplus function.
This soft constraint shifts the solution away from the unconstrained minimum at the origin, as shown in Figure~\ref{fig: groundtruth sample visualization}.
The parameter $\alpha_1 \in \mathbb{R}$ controls the strength of the constraint.
Due to the symmetry of both the objective and the constraint, the optimal solutions lie along a diagonal direction and move along it as $\alpha_1$ varies.

\subsubsection{Data collection}

We consider three problem dimensions $d_x = 100, 150, 200$.
For each case, we uniformly sample $90$ parameter values $\alpha_1 \in [0, 30]$, with $80$ used for training and $10$ for validation.
For each $\alpha_1$, we draw $100$ initial guesses $x^{0}$ uniformly from $[-2, 2]^{d_x}$.
We use vanilla gradient descent as the solver $\pi$, with a fixed step size of $0.1$, and terminate when the residual satisfies $\| x^{m} - x^{m-1} \| \leq 10^{-2}$.
From each solver run, we collect the last $10$ iterates to form the $10$-neighborhood dataset.
In this illustrative example, all converged solutions are treated as optimal.

\subsubsection{Test results}

\begin{figure}[t]
    \centering
    \includegraphics[width=0.85\textwidth]{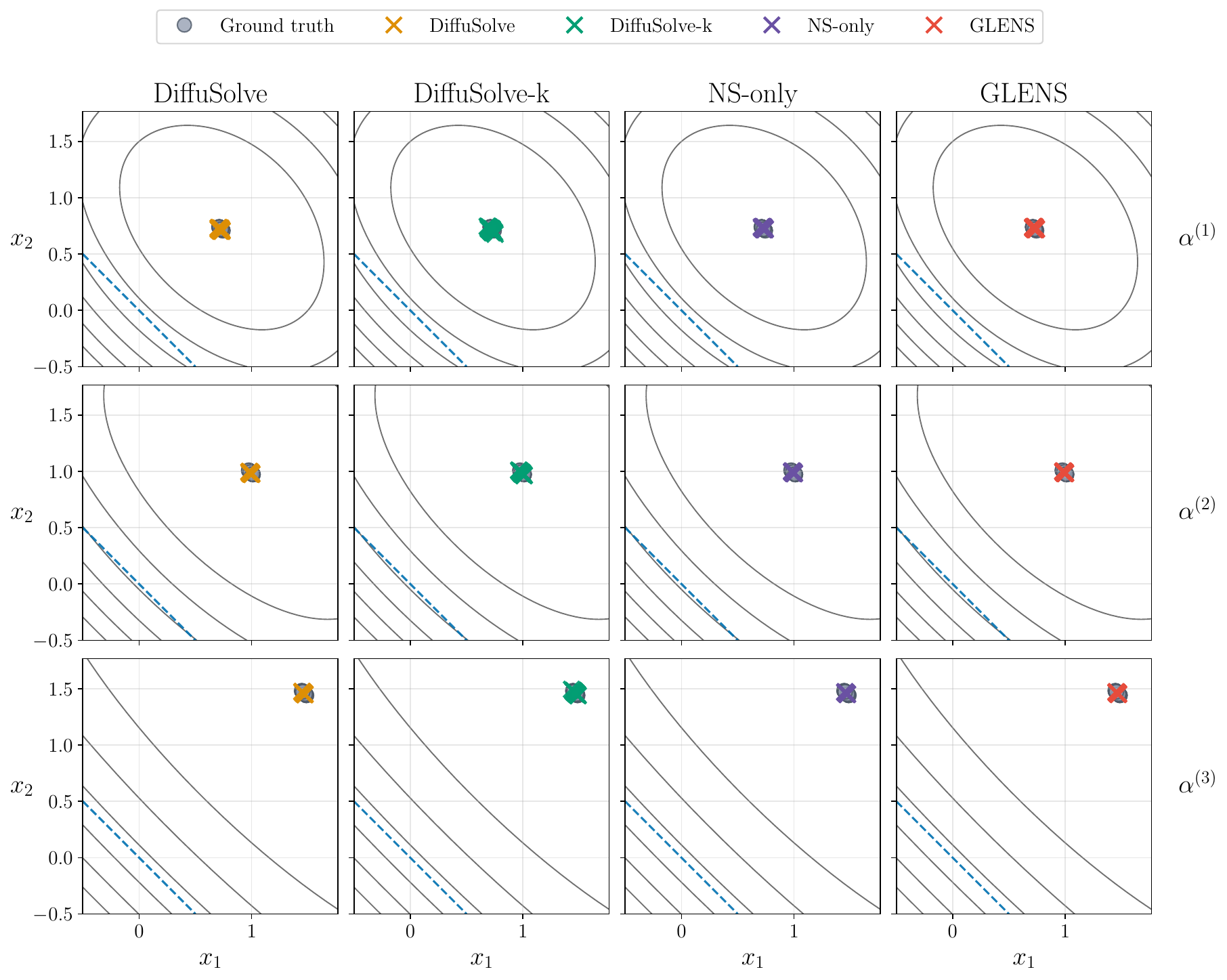}
    \caption{
    First two dimensions of the ground truth optima (grey circles) and sampled initial guesses (colored crosses) for the soft-constrained QP under three unseen test parameters $\alpha^{(1)} = 3.83$, $\alpha^{(2)} = 8.21$, and $\alpha^{(3)} = 28.79$, with each row corresponding to one parameter.
    Each column corresponds to a different method.
    The objective contours are shown in the background, and the linear constraint $x_1 + x_2 = 0$ is indicated by the blue dashed line.
    As $\alpha_1$ increases, the optimal solutions shift upward, and the contours become more distorted.
    All diffusion-based methods capture how solutions vary with respect to constraint weight $\alpha_1$, with DiffuSolve-k exhibiting slightly larger sample variance.
    }
    \label{fig: qp soft constrained, sample visualization}
\end{figure}

At test time, we sample $100$ unseen parameters $\alpha \in [0,50]$ and generate $100$ initial guesses from each method per test parameter.

Figure~\ref{fig: qp soft constrained, sample visualization} shows the first two dimensions of 100 ground truth optima and 100 sampled initial guesses for three unseen test parameters $\alpha^{(1)} = 3.83$, $\alpha^{(2)} = 8.21$, and $\alpha^{(3)} = 28.79$, for problem dimension $d_x=100$.
The ground truth samples, marked by grey circles in every subplot, are obtained by running the solver from uniformly sampled initial guesses until convergence.
Samples from each method are shown as colored crosses.
The objective contours are plotted in the background, and the linear constraint on the first two dimensions, $x_1 + x_2 = 0$, is illustrated by the dashed blue line.
As $\alpha$ increases from $\alpha^{(1)}$ to $\alpha^{(3)}$, the constraint becomes stronger, causing the optimal solutions to shift upward while also distorting the quadratic contours.

In this unimodal example, DiffuSolve serves as a strong baseline and successfully captures how the solution varies with the constraint weight $\alpha$.
This controlled setting shows that, when the solution distribution has a simple structure, diffusion-based global search trained on converged optima can already provide accurate predictions.
The \neighbormodel and \methodname, trained on the $10$-neighborhood dataset, achieve similarly accurate predictions, while DiffuSolve-k, trained on the same dataset without additional modeling, exhibits slightly higher sample variance.
This observation provides a useful reference point for the more challenging non-convex examples later with multimodal structure.

To quantitatively evaluate the quality of sampled initial guesses, we warm-start the gradient descent solver from each sample and record the number of iterations required for convergence, which corresponds to the $k$-neighborhood.
In Table~\ref{tab: qp himmelblau k neighbor results}, we report results for problem dimensions $d_x=100,150,200$, including the empirical cumulative distribution for $k=1,3,6$.
For example, for \methodname with $d_x=100$, the cumulative distribution at $k=1$ is $94.05\%$, indicating that $94.05\%$ of samples already satisfy the solver's termination criterion.
We also report the mean and standard deviation over all $10{,}000$ samples, where a lower mean indicates closer to convergence.
Uniformly sampled initial guesses, which do not exploit solution structure, suffer from the curse of dimensionality: samples are sparse and tend to lie near the boundary of the high-dimensional hypercube, requiring over $40$ iterations on average to converge.
DiffuSolve significantly improves convergence speed, but DiffuSolve-k exhibits degraded performance due to the lack of explicit modeling of neighborhood structure.
Both NS-only and \methodname benefit from the augmented data and modeling of local geometry, achieving the strongest performance and nearly identical results across different problem dimensions.

\subsection{Modified Himmelblau function}

\begin{figure}[t]
    \centering
    \includegraphics[width=0.85\textwidth]{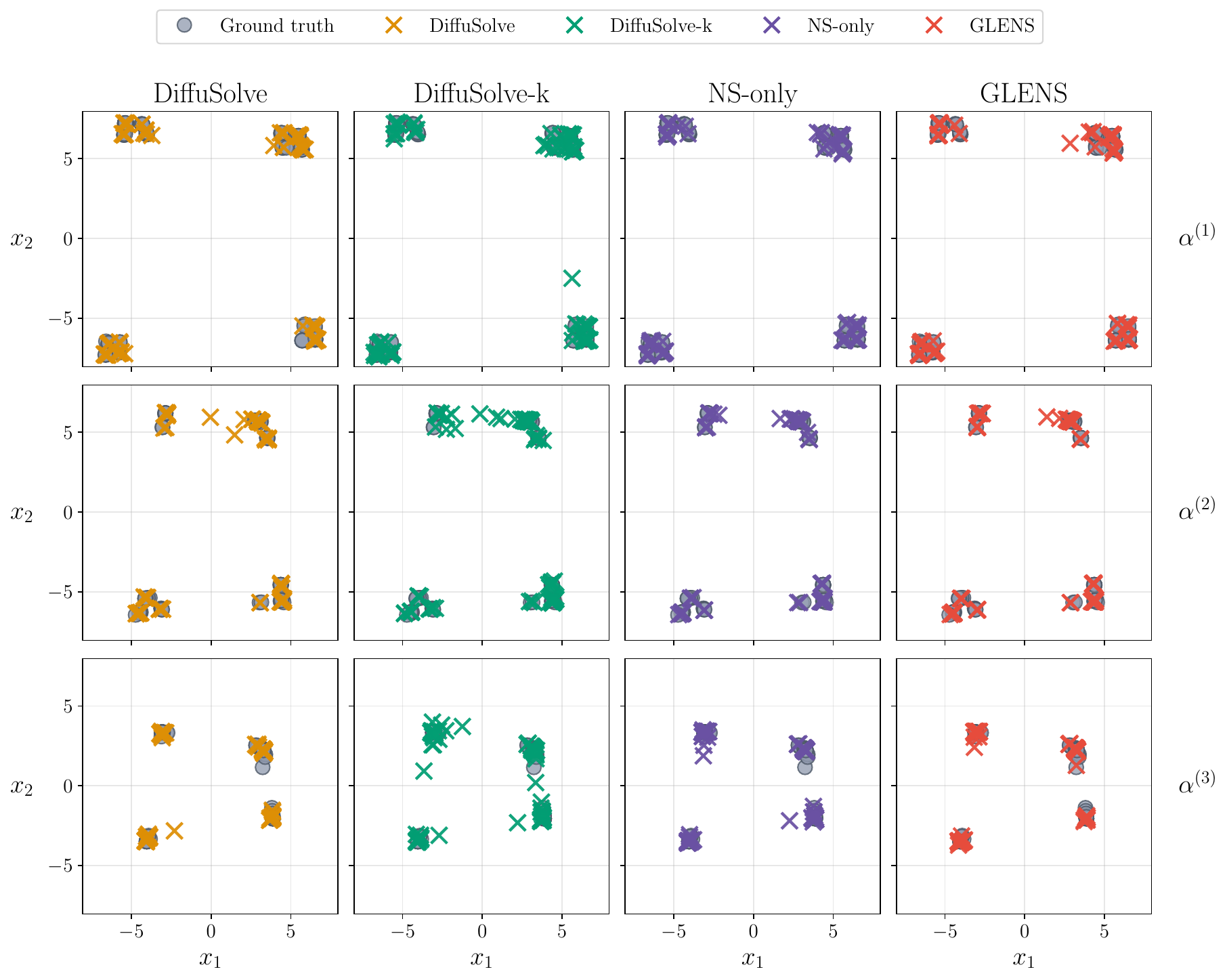}
    \caption{
    First two dimensions of the ground truth optima (grey circles) and sampled initial guesses (colored crosses) for the modified Himmelblau function under three unseen test parameters $\alpha^{(1)} = (36.44, 46.90)$, $\alpha^{(2)} = (14.41, 35.60)$, and $\alpha^{(3)} = (12.90, 8.15)$, with each row corresponding to one parameter.
    Each column corresponds to a different method.
    \methodname produces more concentrated samples around the ground truth optima while preserving the multimodal structure.
    }
    \label{fig: himmelblau, sample visualization}
\end{figure}

\subsubsection{Problem setup}

We consider the following parametric modification of the Himmelblau function~\cite{himmelblau1972applied}, extended to $d_x$ dimensions:
\begin{align}\label{eq: himmelblau objective}
    \min_{x \in \mathbb{R}^{d_x}} \quad
    & \frac{2}{d_x}\sum_{i=1}^{d_x/2}
    \left[
        \Big( x_{2i-1}^2 + x_{2i} - \alpha_1 \Big)^2
        +
        \Big( x_{2i-1} + x_{2i}^2 - \alpha_2 \Big)^2
    \right] \nonumber \\
    & + \lambda \frac{2}{d_x-2} \sum_{i=1}^{d_x/2-1}
    \Big( x_{2i-1} - x_{2i+1} \Big)^2.
\end{align}
When $d_x=2$, $\lambda=0$, $\alpha_1=11$, and $\alpha_2=7$, this reduces to the classical Himmelblau function, a standard non-convex test function with four local optima.
We extend the classical function by summing the Himmelblau structure over pairs of dimensions and adding a soft chained coupling term.
The problem parameter $\alpha = (\alpha_1, \alpha_2) \in \mathbb{R}^2$ controls the locations of the local optima in the solution space.
The coupling weight $\lambda$ is fixed for the chained coupling constraint.
This objective exhibits a highly non-convex landscape with multiple local optima grouped in clusters, as shown in Figure~\ref{fig: groundtruth sample visualization}.

\subsubsection{Data collection}
We consider three different problem dimensions $d_x=100, 150, 200$, and set $\lambda=10$.
For each case, we uniformly sample $90$ different parameter instances $\alpha$ from $[1, 50]^2$, with 80 for training and 10 for validation.
For each $\alpha$, we draw $100$ initial guesses $x^{0}$ uniformly from $[-10, 10]^{d_x}$.
We solve the optimization problem using the L-BFGS-B algorithm~\cite{byrd1995limited} implemented in SciPy~\cite{2020SciPy-NMeth} with tolerance $10^{-3}$, and collect the last 10 iterates prior to convergence as the $10$-neighborhood dataset.

\subsubsection{Test results}

At test time, we sample $100$ unseen parameters $\alpha \in [1, 50]^2$ and generate $100$ initial guesses from each method per test parameter.

Figure~\ref{fig: himmelblau, sample visualization} shows the first two dimensions of ground truth local optima and sampled initial guesses from various methods when problem dimension $d_x=100$ for three unseen test parameters $\alpha^{(1)} = (36.44, 46.90), \alpha^{(2)} = (14.41, 35.60), \alpha^{(3)} = (12.90, 8.15)$, with one $\alpha$ for each row.
The ground truth exhibits distinct local optima grouped in four clusters, and all diffusion-based global search methods capture this multimodal structure.
The samples are visually discriminative for this problem.
DiffuSolve, trained only on $1$-neighborhood data, produces samples with noticeably larger variance for parameter $\alpha^{(2)}$, while DiffuSolve-k exhibits even larger variance across all three test parameters.
In contrast, \methodname generates samples that are more tightly concentrated around the ground truth local optima while maintaining solution diversity.
This suggests that \methodname better captures how the locally optimal solution distribution varies with problem parameters while preserving the multimodal structure.

We also warm-start the L-BFGS-B solver from each sample and report the same empirical cumulative distribution of the $k$-neighborhood and the statistics in Table~\ref{tab: qp himmelblau k neighbor results}, for problem dimensions $d_x=100,150,200$.
The uniform samples still perform poorly and suffer from the curse of dimensionality.
DiffuSolve-k generates samples with significantly lower quality than other diffusion-based methods, as it is trained on the $k$-neighborhood dataset without additional modeling.
The \neighbormodel achieves better average results than DiffuSolve, and the \solvermodel further pushes the samples closer to convergence, such that \methodname generates the most concentrated samples and has the best average quality overall.

\begin{table}[t]
    \caption{$k$-neighborhood results when warm-starting from samples over unseen test parameters.
    $^\dagger$~Solved by gradient descent; 
    $^\ddagger$~solved by L-BFGS-B.
    Higher cumulative distribution ($\uparrow$) and lower $k$-neighborhood means ($\downarrow$) indicate faster convergence.
    }
    \label{tab: qp himmelblau k neighbor results}
    \fontsize{7}{7}\selectfont
    \setlength{\tabcolsep}{2pt}
    \renewcommand{\arraystretch}{0.85}
    \begin{tabular}{lclccc>{\centering\arraybackslash}p{1.2cm}>{\centering\arraybackslash}p{1.2cm}}
    \toprule
    \multirow{2}{*}{Problem} & \multirow{2}{*}{\makecell{Problem \\ Dimension}} & \multirow{2}{*}{Method} &
    \multicolumn{3}{c}{\makecell{Cumulative \\ Distribution \% $\uparrow$}} &
    \multicolumn{2}{c}{\makecell{$k$-neighborhood \\ Stats. $\downarrow$ }} \\
    \cmidrule(lr){4-6}\cmidrule(lr){7-8}
    & & & $k=1$ & $k=3$ & $k=6$ & Mean & Std \\
    \midrule
    
    \multirow{15}{*}{QP$^\dagger$ }
    & \multirow{5}{*}{\makecell[c]{100}}
        & Uniform        & 0.00 & 0.00 &  0.00    & 43.24 & 0.89 \\
    &   & DiffuSolve     & 42.58 & 88.99 & 96.92 & 2.22 & 2.44  \\
    &   & DiffuSolve-k   & 41.47 & 68.92 & 92.29 & 2.94 & 2.76 \\
    &   & NS-only        & 93.92 & 97.06 & 98.59 & \textbf{1.32} & 2.12 \\
    &   & GLENS          & \textbf{94.05} & \textbf{97.06} & \textbf{98.59} & 1.33 & 2.13 \\
    \cmidrule(lr){2-8}
    & \multirow{5}{*}{\makecell[c]{150}}
        & Uniform        & 0.00 & 0.00 &  0.00   & 45.19 & 0.74 \\
    &   & DiffuSolve     & 73.05 & 94.59 & 98.47 & 1.69 &  2.53 \\
    &   & DiffuSolve-k   & 50.33 & 80.35 & 95.15 & 2.48 & 2.85 \\
    &   & NS-only        & 96.39 & 97.52 & 98.94 & 1.30 & 2.25 \\
    &   & GLENS          & \textbf{96.39} & \textbf{97.54} & \textbf{98.94} & \textbf{1.30} & 2.25 \\
    \cmidrule(lr){2-8}
    & \multirow{5}{*}{\makecell[c]{200}}
        & Uniform        & 0.00 & 0.00 & 0.00 & 46.57 & 0.66 \\
    &   & DiffuSolve     & 93.45 & 96.55 & 98.55 & 1.38 & 2.45 \\
    &   & DiffuSolve-k   & 53.70 & 88.61 & 95.12 & 2.24 & 2.86 \\
    &   & NS-only        & \textbf{96.93} & \textbf{98.18} & 98.83 &  1.30 & 2.43 \\
    &   & GLENS          & 96.90 & 98.15 & \textbf{98.83} & \textbf{1.29} & 2.43 \\
    \midrule

    \multirow{15}{*}{Himmelblau$^\ddagger$ }
    & \multirow{5}{*}{\makecell[c]{100}}
        & Uniform        & 0.00 & 0.00 &   0.00   & 25.25 & 6.40  \\
    &   & DiffuSolve     & \textbf{47.30} & 79.45 &   87.27   & 3.04 & 3.90  \\
    &   & DiffuSolve-k   & 5.05 & 36.50 &   60.67   & 6.63 & 5.19  \\
    &   & NS-only        & 26.54 & 85.01 &  91.18    & 2.78 & 3.08  \\
    &   & GLENS          & 40.10 & \textbf{93.55} & \textbf{97.03}     & \textbf{2.04} & 2.16  \\
    \cmidrule(lr){2-8}
    & \multirow{5}{*}{150}
        & Uniform      & 0.00 & 0.00 &   0.00 & 25.61 & 6.47 \\
    &   & DiffuSolve   & 59.39 & 83.94 & 90.03 & 2.58 & 3.53 \\
    &   & DiffuSolve-k  & 7.76 & 46.47 & 67.99 & 5.73 & 4.84  \\
    &   & NS-only      & 50.06 & 87.23 & 93.07 & 2.42 & 3.09  \\
    &   & GLENS        & \textbf{63.79} & \textbf{92.85} & \textbf{96.82} & \textbf{1.88} & 2.63  \\
    \cmidrule(lr){2-8}
    & \multirow{5}{*}{200}
        & Uniform      & 0.00 & 0.00 &   0.00 & 28.33 & 6.88  \\
    &   & DiffuSolve   & 47.98 & 74.37 & 83.57 & 3.50 & 4.48  \\
    &   & DiffuSolve-k   & 1.31 & 24.74 & 50.25 & 7.87 & 5.55  \\
    &   & NS-only      & 43.89 & 82.07 & 89.28 & 2.97 & 3.64  \\
    &   & GLENS        & \textbf{57.48} & \textbf{91.06} & \textbf{95.99} & \textbf{2.15} & 2.80 \\
    \midrule

    \multirow{15}{*}{Rosenbrock$^\ddagger$ }
    & \multirow{5}{*}{\makecell[c]{100}}
        & Uniform        & 0.00 & 0.00 & 0.00 & 30.19 & 14.31 \\
    &   & DiffuSolve     & 0.04 & 75.28 & \textbf{98.19} & 3.39 & 6.71  \\
    &   & DiffuSolve-k   & 8.59 & 49.85 & 75.39 & 6.02 & 12.78  \\
    &   & NS-only        & 37.35 & 87.69 & 96.62 & 3.31 &  11.86 \\
    &   & GLENS          & \textbf{40.00} & \textbf{90.25} & 96.73 & \textbf{3.01} & 10.29  \\
    \cmidrule(lr){2-8}
    & \multirow{5}{*}{150}
         & Uniform        & 0.00 & 0.00 & 0.00 & 31.22 & 11.19  \\
    &   & DiffuSolve     & 63.13 & \textbf{96.73} & \textbf{97.61} & 1.77 & 3.91  \\
    &   & DiffuSolve-k   & 12.40 & 51.79 & 75.86 & 4.50 & 4.89  \\
    &   & NS-only        & 69.52 & 94.82 & 95.86 & 1.78 & 2.60 \\
    &   & GLENS          & \textbf{76.37} & 95.08 & 95.77 & \textbf{1.66} & 2.32 \\
    \cmidrule(lr){2-8}
    & \multirow{5}{*}{200}
         & Uniform        & 0.00 & 0.00 & 0.00 & 32.60 & 11.24 \\
    &   & DiffuSolve     & 86.04 & 96.92 & 97.27 & 1.50 & 2.53 \\
    &   & DiffuSolve-k   & 21.01 & 39.80 & 56.60 & 5.99 & 5.79 \\
    &   & NS-only        & 83.47 & \textbf{99.47} & \textbf{99.62} & \textbf{1.22} & 1.09  \\
    &   & GLENS          & \textbf{86.47} & 98.86 & 98.97 & 1.27 & 1.51 \\
    \midrule

    \multirow{15}{*}{Levy$^\ddagger$ }
    & \multirow{5}{*}{\makecell[c]{100}}
        & Uniform        & 0.00 & 0.01 & 0.03 & 19.35 & 3.05 \\
    &   & DiffuSolve      & 0.04 & 2.76 & 49.04 & 6.88 & 2.51 \\
    &   & DiffuSolve-k    & 0.04 & 0.38 & 11.51 & 10.08 & 3.12 \\
    &   & NS-only         & 0.10 & 6.64 & 41.14 & 7.50 & 3.06 \\
    &   & GLENS           & \textbf{0.17} & \textbf{13.55} & \textbf{67.26} & \textbf{5.79} & 2.46 \\
    \cmidrule(lr){2-8}
    & \multirow{5}{*}{150}
         & Uniform        & 0.00 & 0.00 & 0.01 & 19.61 & 2.57 \\
    &   & DiffuSolve     & 0.04 & 3.25 & 43.14 &  7.22 & 2.49 \\
    &   & DiffuSolve-k   & 0.04 & 0.45 & 9.74 & 9.58 &  2.57 \\
    &   & NS-only        & \textbf{0.11} & 4.25 & 51.68 & 6.67 & 2.46 \\
    &   & GLENS          & 0.06 & \textbf{10.63} & \textbf{78.21} &  \textbf{5.32} & 1.94  \\
    \cmidrule(lr){2-8}
    & \multirow{5}{*}{200}
         & Uniform        & 0.00 & 0.01 & 0.01 & 19.87 & 2.40 \\
    &   & DiffuSolve     & 0.00 & 13.59 & 56.57 & 6.30 & 2.57 \\
    &   & DiffuSolve-k   & 0.06 & 4.15 & 36.10 & 7.46 &  2.54\\
    &   & NS-only        & \textbf{0.12} & 15.97 & 62.76 & 5.85 &  2.40 \\
    &   & GLENS          & 0.05 & \textbf{32.85} & \textbf{82.21} & \textbf{4.67} &  2.02 \\
    \bottomrule
    \end{tabular}
\end{table}
\FloatBarrier

\subsection{Modified Rosenbrock function}

\begin{figure}[t]
    \centering
    \includegraphics[width=0.85\textwidth]{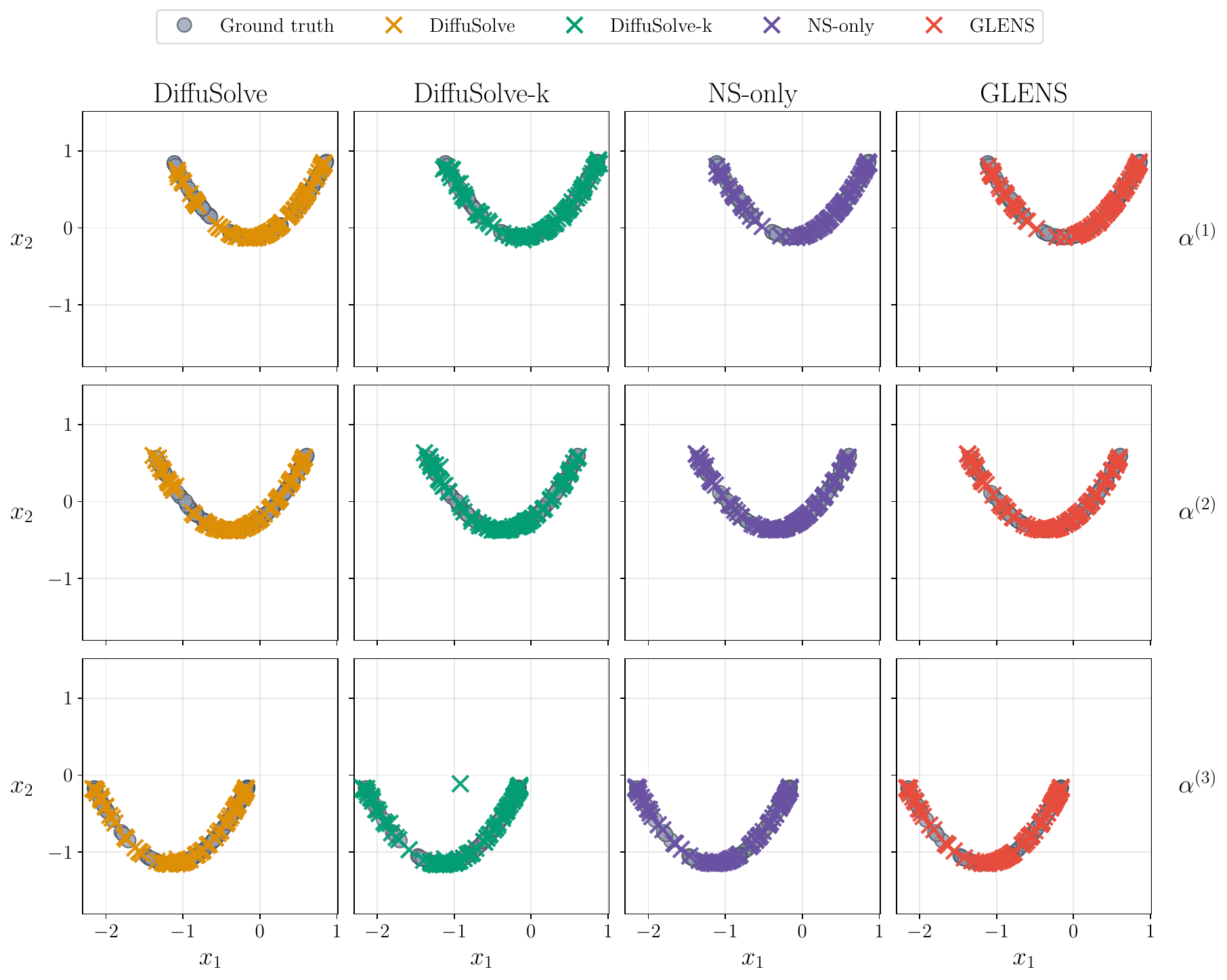}
    \caption{
    First two dimensions of the ground truth optima (grey circles) and sampled initial guesses (colored crosses) for the modified Rosenbrock function under three unseen test parameters $\alpha^{(1)} = (80.23, -0.13)$, $\alpha^{(2)}=(193.97, -0.38)$, and $ \alpha^{(3)} = (91.04, -0.44)$, with each row corresponding to one parameter.
    Each column corresponds to a different method.
    All methods capture the overall solution structure, although DiffuSolve-k exhibits slightly larger variance in its samples.
    }
    \label{fig: rosenbrock, sample visualization}
\end{figure}

\subsubsection{Problem setup}

We consider the following parametric modification of the Rosenbrock function~\cite{rosenbrock1960automatic}, extended to $d_x$ dimensions:
\begin{align}\label{eq: modified rosenbrock}
    \min_{x \in \mathbb{R}^{d_x}} \quad
    & \sum_{i=1}^{d_x/2}
    \left[
       \alpha_1 \Big( (x_{2i-1} - \alpha_2)^2 - (x_{2i} - \alpha_2) \Big)^2
        +
        \Big( (x_{2i-1} - \alpha_2) - 1 \Big)^2
    \right],
\end{align}
where $\alpha_1$ controls the width of the valley and $\alpha_2$ introduces a global shift of the solutions.
When $\alpha_1=100$ and $\alpha_2=0$, this reduces to the extended Rosenbrock function~\cite{dixon1994effect}, a widely used global optimization test function characterized by a long, narrow valley with a curved structure.
The resulting local optima structure in the first two dimensions is shown in Figure~\ref{fig: groundtruth sample visualization}.

\subsubsection{Data collection}

We consider three different problem dimensions $d_x=100, 150, 200$, and uniformly sample $90$ different parameter instances with $\alpha_1 \in [50, 200], \alpha_2 \in [-1.5, 0]$.
Among them, $80$ are used for training and $10$ for validation.
For each $\alpha$, we draw $100$ initial guesses $x^{0}$ uniformly from $[-4, 4]^{d_x}$, and solve with L-BFGS-B algorithm~\cite{byrd1995limited} in SciPy~\cite{2020SciPy-NMeth} with tolerance $10^{-3}$.
We then collect the $10$-neighborhood dataset.

\subsubsection{Test results}

At test time, we sample $100$ unseen parameters $\alpha_1 \in [50, 200]$ and $ \alpha_2 \in [-1.5, 0]$, and generate $100$ initial guesses from each method per test parameter.

In Figure~\ref{fig: rosenbrock, sample visualization}, we visualize the first two dimensions of the ground truth local optima and the sampled initial guesses with problem dimension $d_x=100$, for three test parameters $\alpha^{(1)} = (80.23, -0.13)$, $\alpha^{(2)}=(193.97, -0.38)$, and $ \alpha^{(3)} = (91.04, -0.44)$.
The ground truth samples exhibit a parabolic structure along the valley and shift with the parameter $\alpha_2$.
All diffusion-based methods capture the overall solution structure reasonably well in this example, while DiffuSolve-k, which is trained on the $k$-neighborhood dataset without additional modeling, still shows higher variance compared to other methods.

The $k$-neighborhood statistics in Table~\ref{tab: qp himmelblau k neighbor results} provide a more precise comparison of sample quality.
Across problem dimensions $d_x=100,150,200$, samples from NS-only and \methodname tend to lie in smaller neighborhoods of the optima than DiffuSolve-k and are competitive with or better than DiffuSolve.
For problem dimension $d_x=200$, the \neighbormodel alone achieves over $99\%$ of samples converging within $3$ solver iterations.

\subsection{Modified Levy function}

\begin{figure}[t]
    \centering
    \includegraphics[width=0.85\textwidth]{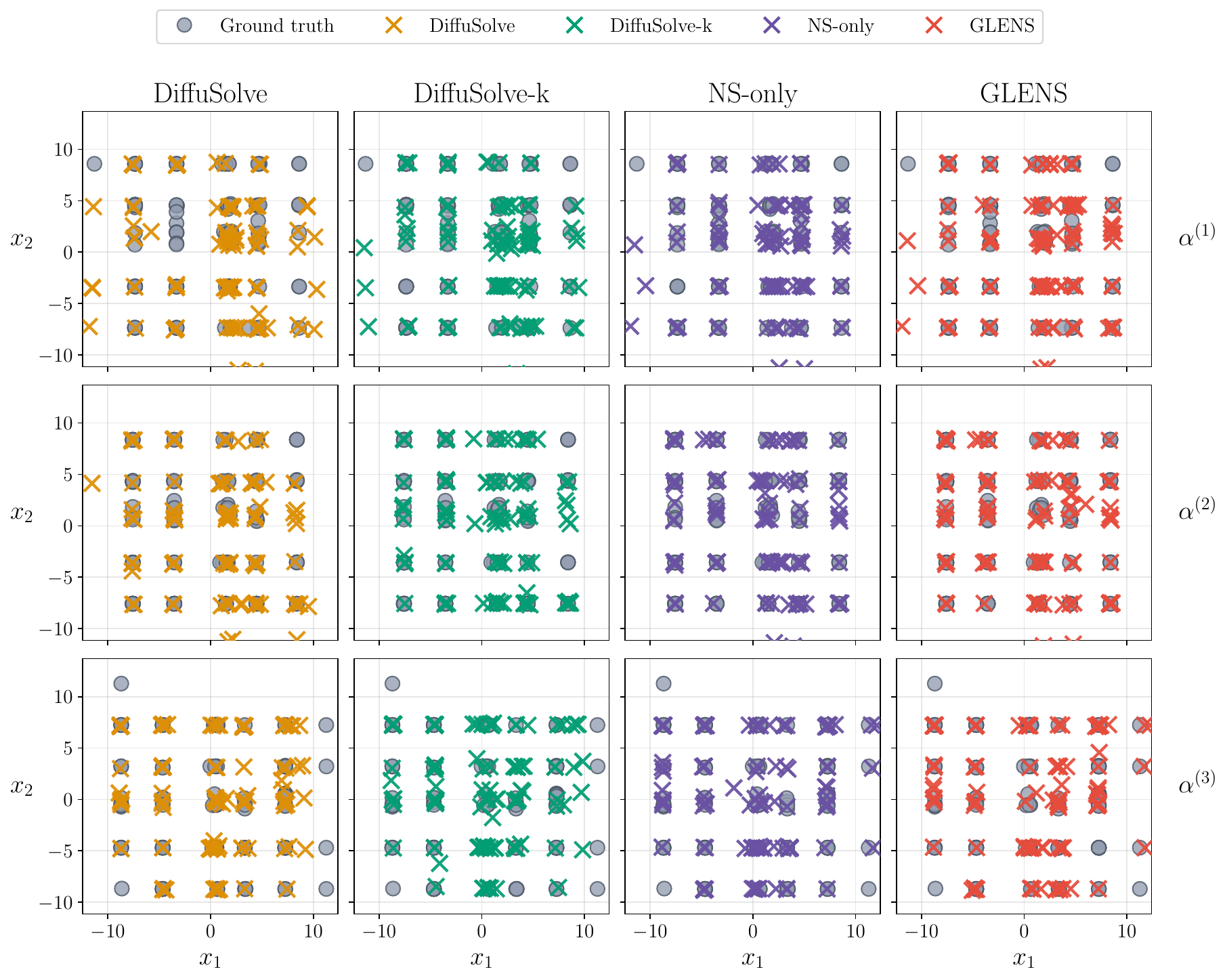}
    \caption{
    First two dimensions of the ground truth optima (grey circles) and sampled initial guesses (colored crosses) for the modified Levy function under three unseen test parameters $\alpha^{(1)} = 0.89$, $\alpha^{(2)}=0.67$, and $ \alpha^{(3)} = -0.45$, with each row corresponding to one parameter.
    Each column corresponds to a different method.
    The ground truth samples exhibit many local optima with highly multimodal structures, making this problem more challenging than previous examples.
    DiffuSolve and DiffuSolve-k fail to predict certain modes accurately.
    In comparison, \methodname better aligns the generated samples with the ground truth locally optimal solution structure across the observed modes.
    }
    \label{fig: levy, sample visualization}
\end{figure}

\subsubsection{Problem setup}

We consider the following parametric modification of the Levy function~\cite{laguna2005experimental}, extended to $d_x$ dimensions.
Let
\begin{align}
    w_i
    \coloneqq
    1 + \frac{x_i - \alpha_1 - 1}{4},
    \quad i=1,\ldots,d_x .
\end{align}
The objective is
\begin{align}
\min_{x \in \mathbb{R}^{d_x}}\;
\sin^2\!\left( \pi w_1 \right) 
& + \sum_{i=1}^{d_x-1}
\left[
    \Big( w_i - 1 \Big)^2
    \left(
        1 + 10 \sin^2\!\left( \pi w_i + 1 \right)
    \right)
\right] 
\nonumber \\
& + \left( w_{d_x} - 1 \right)^2
\left[
    1 + \sin^2\!\left( 2\pi w_{d_x} \right)
\right].
\end{align}
When $\alpha_1=0$, this reduces to the classical Levy function, a multimodal global optimization test function with many local optima.
The parameter $\alpha_1$ introduces a global shift of the solution structure.
The resulting local optima exhibit complex multimodal patterns, as shown in Figure~\ref{fig: groundtruth sample visualization}.

\subsubsection{Data collection}

For problem dimensions $d_x=100, 150$, and $200$, we uniformly sample $90$ parameter instances with $\alpha_1 \in [-2.0, 2.0]$, among which $80$ are used for training and $10$ for validation.
For each parameter, we draw $100$ initial guesses $x^{0}$ uniformly from $[-10, 10]^{d_x}$, and solve the problem using the L-BFGS-B algorithm~\cite{byrd1995limited} implemented in SciPy~\cite{2020SciPy-NMeth} with tolerance $10^{-3}$.
We collect the last $10$ iterates before convergence as the $10$-neighborhood dataset.

\subsubsection{Test results}

At test time, we sample $100$ unseen parameters $\alpha_1 \in [-2.0, 2.0]$, and generate $100$ initial guesses from each method per test parameter.

In Figure~\ref{fig: levy, sample visualization}, we visualize the first two dimensions of $100$ ground truth local optima together with $100$ sampled initial guesses for three test parameters $\alpha^{(1)} = 0.89$, $\alpha^{(2)}=0.67$, and $ \alpha^{(3)} = -0.45$, with problem dimension $d_x=100$.
The ground truth samples exhibit many local optima with complex structures.
All methods capture the overall solution pattern to some extent, but generally exhibit higher noise compared to previous examples.
DiffuSolve finds it challenging to predict certain modes, while DiffuSolve-k shows even larger local variance.
\methodname produces higher-quality samples with more accurate predictions than other methods and preserves the diversity of the solutions.

The difficulty of this problem is also reflected in the statistics in Table~\ref{tab: qp himmelblau k neighbor results}, where the gap between the Uniform method and the learning-based methods is smaller than in the previous examples.
Consistent with the qualitative visualization, the $k$-neighborhood statistics show that \methodname produces samples closer to convergence, especially for $k=3$ and $k=6$.
In this setting, the \solvermodel plays a more important role because learning the solution distribution is challenging for diffusion models alone, enabling \methodname to produce sharper samples that are closer to convergence, while maintaining multimodal coverage.

\section{Hyperparameter analysis}\label{sec: hyperparameter}

To systematically examine the performance and robustness of the proposed method \methodname under different hyperparameter choices, we conduct a series of analyses on the numerical problems introduced earlier.
In particular, we study the effect of
(i) the neighborhood size $k$ in training data,
(ii) the model capacity of \methodname, and
(iii) the solver refinement guidance weight $s_{\text{SB}}$ from the \solvermodel.

For each experiment, we follow the evaluation settings described in the previous section and assess the quality of $10{,}000$  samples across $100$ unseen test parameters for problem dimension $d_x=200$, by warm-starting the corresponding numerical solver.
We report the cumulative distribution of the resulting $k$-neighborhoods, for $k=1,3,6$, over all samples, where the $k$-neighborhood corresponds to the number of solver iterations required to reach convergence from a given initial guess.
We also summarize the mean and standard deviation of the observed $k$-neighborhood values.

\subsection{Neighborhood size $k$}

\begin{table}[t]
    \caption{Effect of using different neighborhood size $k$ in training data for \methodname.
    $k=1$ indicates the DiffuSolve results.
    $^\dagger$~Solved by gradient descent; 
    $^\ddagger$~solved by L-BFGS-B.
    Higher cumulative distribution ($\uparrow$) and lower $k$-neighborhood means ($\downarrow$) indicate faster convergence.
    }
    \label{tab:different_k_neighborhood}
    \begin{tabular}{lcccc>{\centering\arraybackslash}p{1.2cm}>{\centering\arraybackslash}p{1.2cm}}
    \toprule
    \multirow{2}{*}{Problem} & \multirow{2}{*}
    {\makecell{Training \\ $k$-neighborhood \\ dataset}} &
    \multicolumn{3}{c}{\makecell{Cumulative \\ Distribution \% $\uparrow$}} &
    \multicolumn{2}{c}{\makecell{$k$-neighborhood \\ Stats. $\downarrow$ }} \\
    \cmidrule(lr){3-5}\cmidrule(lr){6-7}
    & & $k=1$ & $k=3$ & $k=6$ & Mean & Std \\
    \midrule
    
    \multirow{4}{*}{QP$^\dagger$}
    & 1  & 93.45 & 96.55 & 98.55 & 1.38 & 2.45 \\
    & 5  & 91.67 & 94.23 & 96.68 & 1.53 & 2.63 \\
    & 10 & \textbf{96.90} & \textbf{98.15}  & \textbf{98.83} & \textbf{1.29} & 2.43 \\
    & 15 & 88.87 & 97.05 & 98.76 & 1.42 & 2.47 \\
    \midrule
    
    \multirow{4}{*}{Himmelblau$^\ddagger$}
    & 1  & 47.98 & 74.37 & 83.57 & 3.50 & 4.48 \\
    & 5  & \textbf{70.01} & 88.16 & 94.19 & \textbf{2.13} & 3.03 \\
    & 10 & 57.48 & \textbf{91.06} & \textbf{95.99} & 2.15 & 2.80 \\
    & 15 & 56.60 & 88.18 & 95.67 & 2.18 & 2.67 \\
    \midrule
    
    \multirow{4}{*}{Rosenbrock$^\ddagger$}
    & 1  & 86.04 & 96.92 & 97.27 & 1.50 & 2.53 \\
    & 5  & 26.75 & 92.53 & 96.06 & 2.56 & 4.49 \\
    & 10 & \textbf{86.47} & \textbf{98.86} & \textbf{98.97} & \textbf{1.27} & 1.51\\
    & 15 & 82.99 & 94.39 & 95.51 & 1.69 & 3.98 \\
    \midrule
    
    \multirow{4}{*}{Levy$^\ddagger$}
    & 1  & 0.00 & 13.59 & 56.57 & 6.30 & 2.57 \\
    & 5  & 0.02 & 22.98 & 68.11 & 4.97 & 2.01 \\
    & 10 & 0.05 & \textbf{32.85} & \textbf{82.21} & \textbf{4.67} &  2.02\\
    & 15 & \textbf{0.12} & 15.19 & 65.73 & 5.80 & 2.38 \\
    
    \bottomrule
    \end{tabular}
\end{table}

The neighborhood size $k$ controls a trade-off between data coverage and quality.
Larger $k$ provides more free training data and reveals richer neighborhood structure around locally optimal solutions, but may include solver iterates far from the optima, introducing noise and potentially degrading model performance.

Table~\ref{tab:different_k_neighborhood} reports the performance of \methodname trained on different neighborhood sizes $k$, where $k=1$ corresponds to the DiffuSolve baseline.
Incorporating neighborhood data consistently helps: $k=5$ and $k=10$ give significant gains over $k=1$.
However, the effective neighborhood around each optimum is not unlimited, and increasing $k$ to $15$ introduces iterates far from the optima that add noise to the training data and slightly degrade performance.
Across the four problems, $k=10$ provides a good balance between data augmentation and quality, suggesting that the solver's convergence behavior is a useful guideline for how far its iterates remain informative of the local solution landscape.

\subsection{Model capacities}

\begin{table}[t]
    \caption{Effect of different model dimensions for \methodname.
    DiffuSolve only has an ``NS'' dimension, with ``SB'' marked as N/A.
    $^\dagger$~Solved by gradient descent; 
    $^\ddagger$~solved by L-BFGS-B.
    Higher cumulative distribution ($\uparrow$) and lower $k$-neighborhood means ($\downarrow$) indicate faster convergence.
    }
    \label{tab:different_model_size}
    \begin{tabular}{lccccccc>{\centering\arraybackslash}p{1.2cm}>{\centering\arraybackslash}p{1.2cm}}
    \toprule
    \multirow{3}{*}{Problem} &
    \multicolumn{2}{c}{\makecell{Model \\ dim}} &
    \multirow{3}{*}{\makecell{Training \\ Neighbor.}} &
    \multicolumn{3}{c}{\makecell{Cumulative \\ Distribution \% $\uparrow$}} &
    \multicolumn{2}{c}{\makecell{$k$-neighborhood \\ Stats $\downarrow$ }} \\
    \cmidrule(lr){2-3}\cmidrule(lr){5-7}\cmidrule(lr){8-9}
    & NS & SB & & $k=1$ & $k=3$ & $k=6$ & Mean & Std \\
    \midrule
    
    \multirow{12}{*}{QP$^\dagger$}
    & \multirow{3}{*}{32}
        & N/A & 1  & 48.41 & 80.90 & 95.13 & 2.48 & 2.79 \\
    &   & 32  & 10 & 96.14 & 97.73 & 98.85 & 1.30 & 2.34 \\
    &   & 64  & 10 & \textbf{96.18} & \textbf{97.86} & \textbf{98.88} & \textbf{1.30} & 2.33 \\
    \cmidrule(lr){2-9}
    & \multirow{2}{*}{64}
        & N/A & 1 & 93.45 & 96.55 & 98.55 & 1.38 & 2.45\\
    &   & 64  & 10 & 96.90 & 98.15 & \textbf{98.83} & \textbf{1.29} & 2.43 \\
    \cmidrule(lr){2-9}
    & \multirow{3}{*}{128}
        & N/A & 1  & 69.34 & 79.26 & 83.45 & 3.50 & 5.19 \\
    &   & 64  & 10 & \textbf{68.49} & \textbf{89.19} &
    \textbf{94.76} & \textbf{2.14} & 3.27 \\
    &   & 128 & 10 & 68.45 & 89.13 & 94.71 & 2.14 & 3.27 \\
    \midrule
    
    \multirow{12}{*}{Himmelblau$^\ddagger$}
    & \multirow{3}{*}{32}
        & N/A & 1  & 54.50 & 71.24 & 81.18 & 3.74 & 4.83 \\
    &   & 32  & 10 & 52.77 & 86.29 & 93.99 & 2.42 & 3.31 \\
    &   & 64  & 10 & \textbf{54.80} & \textbf{89.56} & \textbf{95.11} & \textbf{2.21} & 3.08 \\
    \cmidrule(lr){2-9}
    & \multirow{2}{*}{64}
        & N/A & 1  & 47.98 & 74.37 & 83.57 & 3.50 & 4.48 \\
    &   & 64  & 10 & \textbf{57.48} & \textbf{91.06} & \textbf{95.99} & \textbf{2.15} & 2.80 \\
    \cmidrule(lr){2-9}
    & \multirow{3}{*}{128}
        & N/A & 1  & 40.71 & 73.73 & 83.85 & 3.65 & 4.58 \\
    &   & 64  & 10 & \textbf{43.14} & \textbf{88.05} & 94.77 & \textbf{2.44} & 3.06     \\
    &   & 128 & 10 & 7.67  & 83.34 & \textbf{96.40} & 2.95 & 2.34 \\
    \midrule
    
    \multirow{12}{*}{Rosenbrock$^\ddagger$}
    & \multirow{3}{*}{32}
        & N/A & 1  & 40.33 & 96.25 & 97.28 & 1.99 & 2.35 \\
    &   & 32  & 10 & \textbf{54.93} & \textbf{95.94} & \textbf{98.03} & \textbf{1.88} & 3.84 \\
    &   & 64  & 10 & 53.73 & 95.63 & 97.87 & 1.91 & 3.87 \\
    \cmidrule(lr){2-9}
    & \multirow{2}{*}{64}
        & N/A & 1  & 86.04 & 96.92 & 97.27 & 1.50 & 2.53 \\
    &   & 64  & 10 & \textbf{86.47} & \textbf{98.86} & \textbf{98.97} & \textbf{1.27} & 1.51 \\
    \cmidrule(lr){2-9}
    & \multirow{3}{*}{128}
        & N/A & 1  & 27.38 & 91.17 & 96.98 & 2.44 & 4.32 \\
    &   & 64  & 10 & \textbf{41.79} & \textbf{91.71} & \textbf{94.86} & \textbf{2.41} & 4.54   \\
    &   & 128 & 10 & 40.23 & 91.23 & 94.90 & 2.43 & 3.95 \\
    \midrule
    
    \multirow{12}{*}{Levy$^\ddagger$}
    & \multirow{3}{*}{32}
        & N/A & 1  & \textbf{0.08} & 0.72 & 10.15 & 10.59 & 3.02 \\
    &   & 32  & 10 & 0.04 & 2.04 & 28.22 & 8.97 &  3.38 \\
    &   & 64  & 10 & 0.01 & \textbf{3.01} & \textbf{40.87} & \textbf{7.55} & 2.78 \\
    \cmidrule(lr){2-9}
    & \multirow{2}{*}{64}
        & N/A & 1  & 0.00 & 13.59 & 56.57 & 6.30 & 2.57 \\
    &   & 64  & 10 & \textbf{0.05} & \textbf{32.85} & \textbf{82.21} & \textbf{4.67} &  2.02 \\
    \cmidrule(lr){2-9}
    & \multirow{3}{*}{128}
        & N/A & 1  & \textbf{0.03} & 7.53 & 43.91 & 7.10 & 2.67 \\
    &   & 64  & 10 & 0.02 & \textbf{18.24} & \textbf{75.18} & \textbf{5.36} &  2.24  \\
    &   & 128 & 10 & 0.01 & 6.23 & 36.48 & 7.80 & 2.98 \\
    \bottomrule
    \end{tabular}
\end{table}

Both the \neighbormodel and the \solvermodel use a U-Net backbone, whose base feature dimension controls the model capacity.
Table~\ref{tab:different_model_size} evaluates \methodname under U-Net dimensions $32$, $64$, and $128$, alongside DiffuSolve at the same dimensions for comparison.

Increasing the dimension from $32$ to $64$ consistently improves performance, indicating that a minimum capacity is needed to capture the multimodal solution structure.
Further increasing the dimension to $128$ tends to degrade performance.
The \solvermodel appears especially prone to overfitting: results often improve when its dimension is reduced from $128$ to $64$, suggesting that modeling a larger $k$-neighborhood distribution requires a more balanced capacity.
A base dimension of $64$ provides a good trade-off between expressiveness and stability across all four problems.
Across model dimensions, \methodname consistently outperforms the baselines.

\subsection{Guidance weight for \solvermodel}

\begin{table}[t]
    \caption{Effect of different guidance weight for the \solvermodel in \methodname.
     $^\dagger$~Solved by gradient descent; 
    $^\ddagger$~solved by L-BFGS-B.
    Higher cumulative distribution ($\uparrow$) and lower $k$-neighborhood means ($\downarrow$) indicate faster convergence.
    }
    \label{tab:different_guidance_weight}
    \begin{tabular}{lcccc>{\centering\arraybackslash}p{1.2cm}>{\centering\arraybackslash}p{1.2cm}}
    \toprule
    \multirow{2}{*}{Problem} & \multirow{2}{*}{\makecell{Guidance \\ weight $s_{\text{SB}}$}} &
    \multicolumn{3}{c}{\makecell{Cumulative \\ Distribution \% $\uparrow$}} &
    \multicolumn{2}{c}{\makecell{$k$-neighborhood \\ Stats $\downarrow$ }} \\
    \cmidrule(lr){3-5}\cmidrule(lr){6-7}
    & & $k=1$ & $k=3$ & $k=6$ & Mean & Std \\
    \midrule
    
    \multirow{4}{*}{QP$^\dagger$}
    & 10  & 96.92 & \textbf{98.19} & \textbf{98.83} & \textbf{1.29} & 2.43 \\
    & 50  & \textbf{96.92} & 98.18 & 98.83 & 1.29 & 2.43 \\
    & 100 & 96.90 & 98.15 & 98.83 & 1.29 & 2.43 \\
    & 200 & 96.91 & 98.12 & 98.83 & 1.29 & 2.43 \\
    \midrule
    
    \multirow{4}{*}{Himmelblau$^\ddagger$}
    & 10  & 34.14 & 80.38 & 91.88 & 2.92 & 3.21 \\
    & 50  & 48.15 & 89.49 & \textbf{95.88} & 2.26 & 2.72 \\
    & 100 & \textbf{53.98} & \textbf{90.53} & 95.78 & \textbf{2.17} & 2.85 \\
    & 200 & 0.00 & 0.04 & 15.61 & 11.32 & 5.09 \\
    \midrule
    
    \multirow{4}{*}{Rosenbrock$^\ddagger$}
    & 10  & 83.85 & \textbf{99.35} & \textbf{99.49} & \textbf{1.23} & 1.07 \\
    & 50  & 85.21 & 99.16 & 99.28 & 1.24 & 1.26 \\
    & 100 & 86.47 & 98.86 & 98.97 & 1.27 & 1.51 \\
    & 200 & \textbf{87.58} & 98.14 & 98.40 & 1.35 & 3.33 \\
    \midrule
    
    \multirow{4}{*}{Levy$^\ddagger$}
    & 10  & 0.14 & 28.43 & 78.58 & 4.89 & 2.14 \\
    & 50  & \textbf{0.18} & \textbf{34.36} & \textbf{83.73} & \textbf{4.54} & 1.97 \\
    & 100 & 0.05 & 32.85 & 82.21 & 4.67 &  2.02 \\
    & 200 & 0.04 & 18.67 & 72.22 & 5.45 & 2.12 \\
    
    \bottomrule
    \end{tabular}
\end{table}

As described in Algorithm~\ref{alg:glens_sample}, solver refinement from the \solvermodel is incorporated into diffusion sampling via a classifier guidance step with guidance weight $s_{\text{SB}}$.
In Table~\ref{tab:different_guidance_weight}, we study the effect of varying $s_{\text{SB}}$ while keeping the \neighbormodel fixed.

For the soft-constrained QP and the modified Rosenbrock example, performance is quite consistent regardless of the choice of guidance weight, suggesting that the \neighbormodel already provides high-quality initial guesses in this setting and leaves limited room for further refinement.
For the modified Himmelblau and Levy examples, where the solution structure is more challenging to model, increasing $s_{\text{SB}}$ significantly improves performance up to a moderate level, indicating that the \solvermodel plays a crucial role in refining samples in more complex multimodal landscapes.
However, overly strong guidance with $s_{\text{SB}} = 200$ leads to performance degradation, likely due to overshooting during the refinement process.

Overall, the \solvermodel makes a complementary contribution alongside the \neighbormodel, particularly on harder problems, and a moderate guidance weight $s_{\text{SB}} = 100$ strikes a good balance between refinement strength and stability.

\section{Real-world example: two-robot navigation}\label{sec: real world examples}

\subsection{Problem setup}

We consider a real-world two-robot navigation problem in which two robots aim to reach their respective goal regions in minimum time while avoiding multiple obstacles and colliding with each other, inspired by~\cite{li2025diffusolve}.
This poses the following open-loop trajectory optimization problem:
\begin{align}\label{eq: robot trajectory optimization prob}
\min_{t_f, u(\cdot)} \quad & t_f \nonumber \\
\text{s.t.} \quad
& \dot{\zeta}_j(t) = f(\zeta_j(t), u_j(t)),
&& \forall t \in [0, t_f],\ \forall j \in \{1, 2\}, \nonumber\\
& \zeta_j(0) = \zeta_{\text{init}_j},
&& \forall j \in \{1, 2\}, \nonumber\\
& \|p_j(t_f) - p_{\text{goal}_j}\|_2 \le \varepsilon_{\text{goal}},
&& \forall j \in \{1, 2\}, \nonumber\\
& \|p_j(t) - p_{\text{obs}_i}\|_2 \ge r_{\text{obs}_i} + r_{\text{robot}},
&& \forall t \in [0, t_f],\ i = 1, \ldots, N_{\text{obs}},\ \forall j \in \{1, 2\}, \nonumber\\
& \|p_1(t) - p_2(t)\|_2 \ge 2 r_{\text{robot}},
&& \forall t \in [0, t_f], \nonumber\\
& u_{\min} \le u_j(t) \le u_{\max},
&& \forall t \in [0, t_f],\ \forall j \in \{1, 2\}, \nonumber\\
& t_{f_{\min}} \le t_f \le t_{f_{\max}}.
&&
\end{align}

Here $t_f$ is the terminal time, with bounds $t_{f_{\min}}$ and $t_{f_{\max}}$.
For each robot $j \in \{1,2\}$, $\zeta_j$ is the state, $p_j$ is its position component, and $u_j$ is the control input bounded by $u_{\min}$ and $u_{\max}$.
Each robot starts from a given initial state $\zeta_{\text{init}_j}$ and must reach its goal position $p_{\text{goal}_j}$ within tolerance $\varepsilon_{\text{goal}}$.
The environment contains $N_{\text{obs}}$ circular obstacles with centers $p_{\text{obs}_i}$ and radii $r_{\text{obs}_i}$; the robots have radius $r_{\text{robot}}$.
Obstacle-avoidance and inter-robot separation constraints enforce a minimum distance of $r_{\text{obs}_i} + r_{\text{robot}}$ to each obstacle and $2 r_{\text{robot}}$ between the two robots.

We assume each robot follows the same nonlinear dynamics $f$,
\begin{align}\label{eq: robot dynamics}
    \dot p_x = v \cos\theta, \quad
    \dot p_y = v \sin\theta, \quad
    \dot v = a, \quad
    \dot \theta = \omega,
\end{align}
with state $\zeta_j = (p_{x_j}, p_{y_j}, v_j, \theta_j)$, where $p_j = (p_{x_j}, p_{y_j})$ is the position, $v_j$ is the linear speed, and $\theta_j$ is the orientation.
The control input is $u_j(t) = (a_j, \omega_j)$, consisting of linear acceleration and angular velocity.

To solve problem~\eqref{eq: robot trajectory optimization prob} under the dynamics~\eqref{eq: robot dynamics}, we use a forward shooting method for control transcription.
The time horizon $[0, t_f]$ is discretized into $T$ uniform intervals, and the dynamics are integrated using a fourth-order Runge-Kutta (RK4) method.
The resulting decision variables consist of the terminal time $t_f$ and the control sequence $\{u_{j,k}\}_{k=0}^{T-1}$ for $j = 1, 2$, with goal-reaching, obstacle-avoidance, and robot-separation constraints enforced on the discretized trajectory.

For this problem family, the problem parameter $\alpha$ collects the obstacle positions $p_{\text{obs}_i}$ and radii $r_{\text{obs}_i}$, so that varying $\alpha$ yields different environments in which to search for diverse trajectories.

\subsection{Data collection}

We fix the initial states as
$
\zeta_{\text{init}_1} = (0, 10, 0, 0), \
\zeta_{\text{init}_2} = (10, 10, 0, 0),
$
and set the goal positions to
$
p_{\text{goal}_1} = (10, 0),
\
p_{\text{goal}_2} = (0, 0).
$
The goal tolerance is $\varepsilon_{\text{goal}} = 0.2$, the robot radius is
$r_{\text{robot}} = 0.2$, and the number of obstacles is $N_{\text{obs}} = 2$.
The control bounds are set as
$
a_{\min} = -1,\  a_{\max} = 1,\
\omega_{\min} = -1,\ \omega_{\max} = 1,
$
and the terminal time bound is
$t_{f_{\min}} = 0$ and $t_{f_{\max}} = 20$.

To collect training data, we uniformly sample $90$ obstacle configurations $\alpha$, where obstacle centers $p_{\text{obs}_i}$ are drawn from $[2.0,8.0]^2$ and radii $r_{\text{obs}_i}$ from $[0.5,1.5]$, with 80 for training and 10 for validation.
For each parameter instance, we generate $100$ uniformly sampled initial guesses
by sampling the control sequences with
$a_k \in [-1,1]$, $\omega_k \in [-1,1]$,
and terminal time $t_f \in [0,20]$.
The time horizon is discretized with $T=20$.

\begin{figure}[t]
\centering
\includegraphics[width=1.0\textwidth]{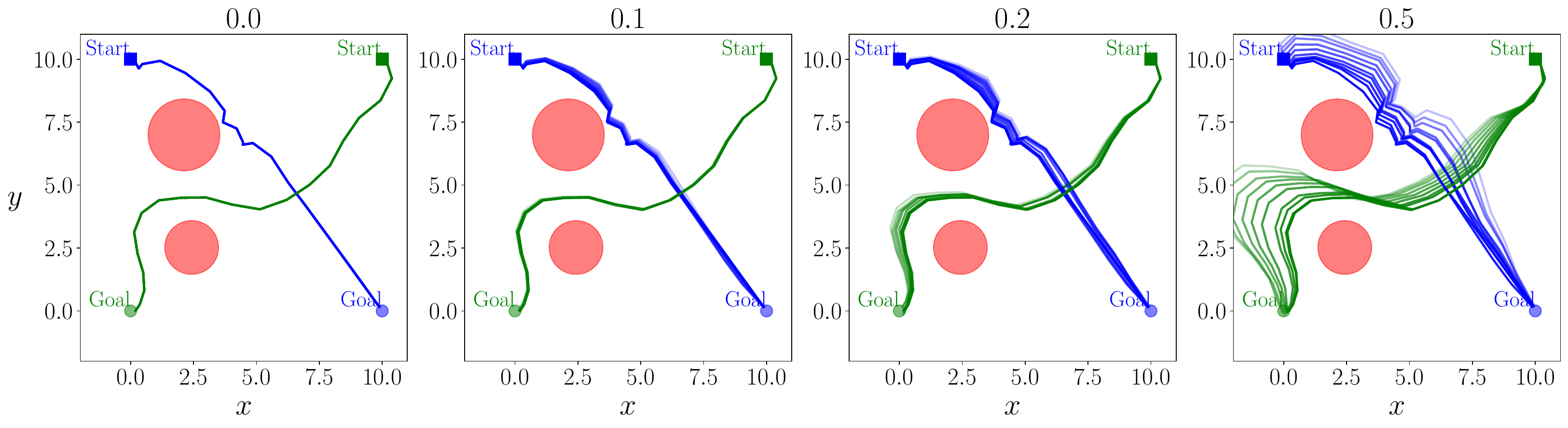}
\caption{
Effect of the filtering threshold on the resulting $k$-neighborhood iterates in the two-robot navigation example.
A low threshold introduces many near-duplicate iterates in the training data, while a large threshold may include iterates that are far from the optimal solutions.
}
\label{fig: robot data filtering}
\end{figure}

Each instance is solved using SNOPT~\cite{gill2005snopt} via pyOptSparse~\cite{wu2020pyoptsparse}, with a maximum major iteration limit of $1000$ and feasibility and optimality tolerances set to $10^{-2}$.
We only keep those solutions that are verified as local optima.
In addition, we observe that the final SNOPT iterates near convergence are often nearly identical, so we apply a simple filtering rule that an iterate is only kept if its distance from the last selected iterate exceeds a predefined threshold.
Figure~\ref{fig: robot data filtering} visualizes the final $10$ iterates obtained after filtering under different thresholds.
We select a threshold of $0.2$ and collect $10$ filtered iterates as the 10-neighborhood data, which provides sufficient variation while remaining close to the optima. 
The resulting dataset is split into training, validation, and test sets, containing $80$, $10$, and $20$ parameter instances, respectively.

\subsection{Test results}

\begin{figure}[t]
\centering
\includegraphics[width=0.7\textwidth]{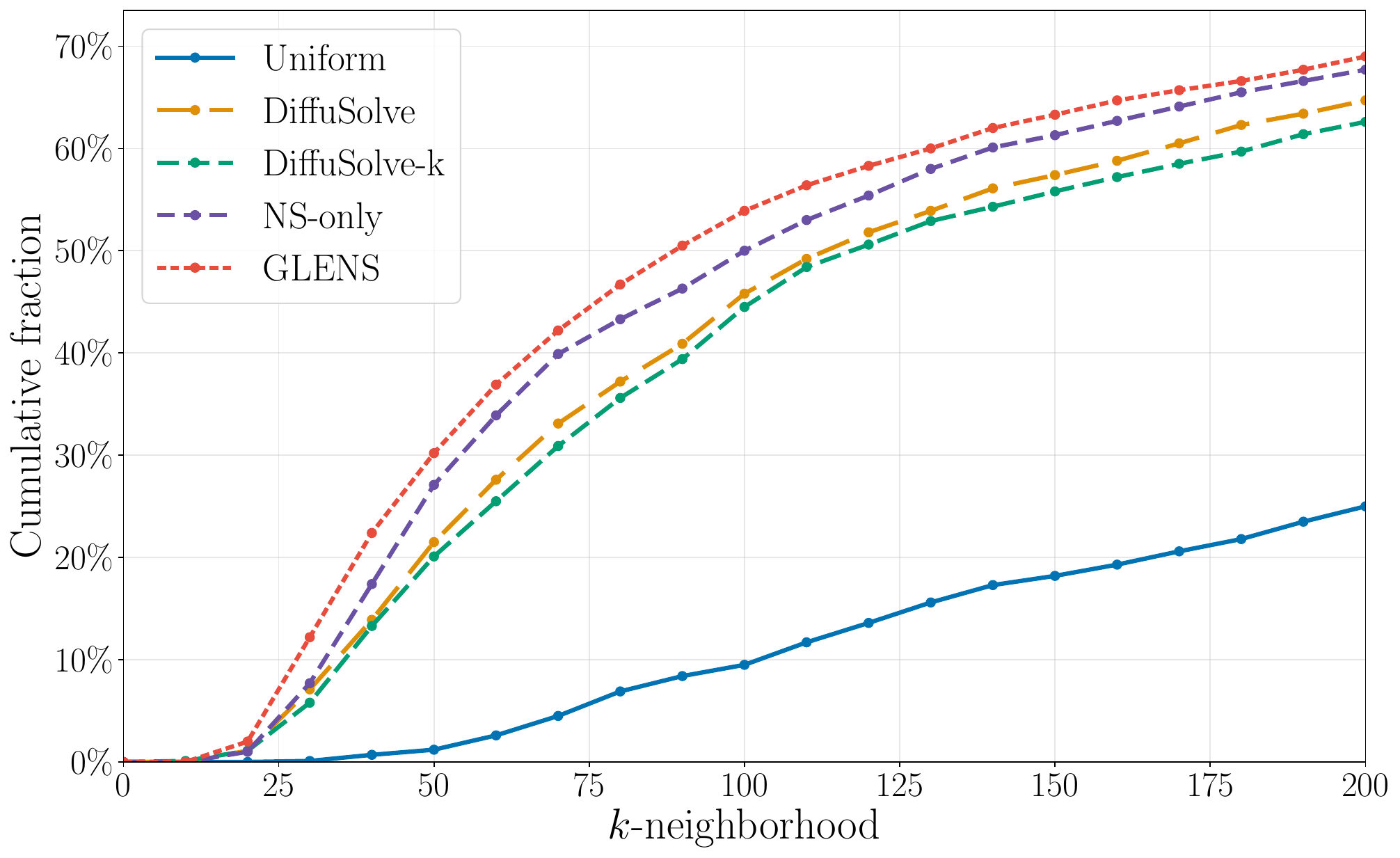}
\caption{
Cumulative distribution of the $k$-neighborhoods reached by sampled initial guesses in the two-robot navigation example.
Steeper curves correspond to samples that lie closer to the converged optima and converge in fewer solver iterations.
Samples generated by \methodname are concentrated in smaller $k$-neighborhoods than those from the baselines.
}
\label{fig: robot, k neighbor cdf}
\end{figure}

\begin{figure}[t]
\centering
\includegraphics[width=0.9\textwidth]{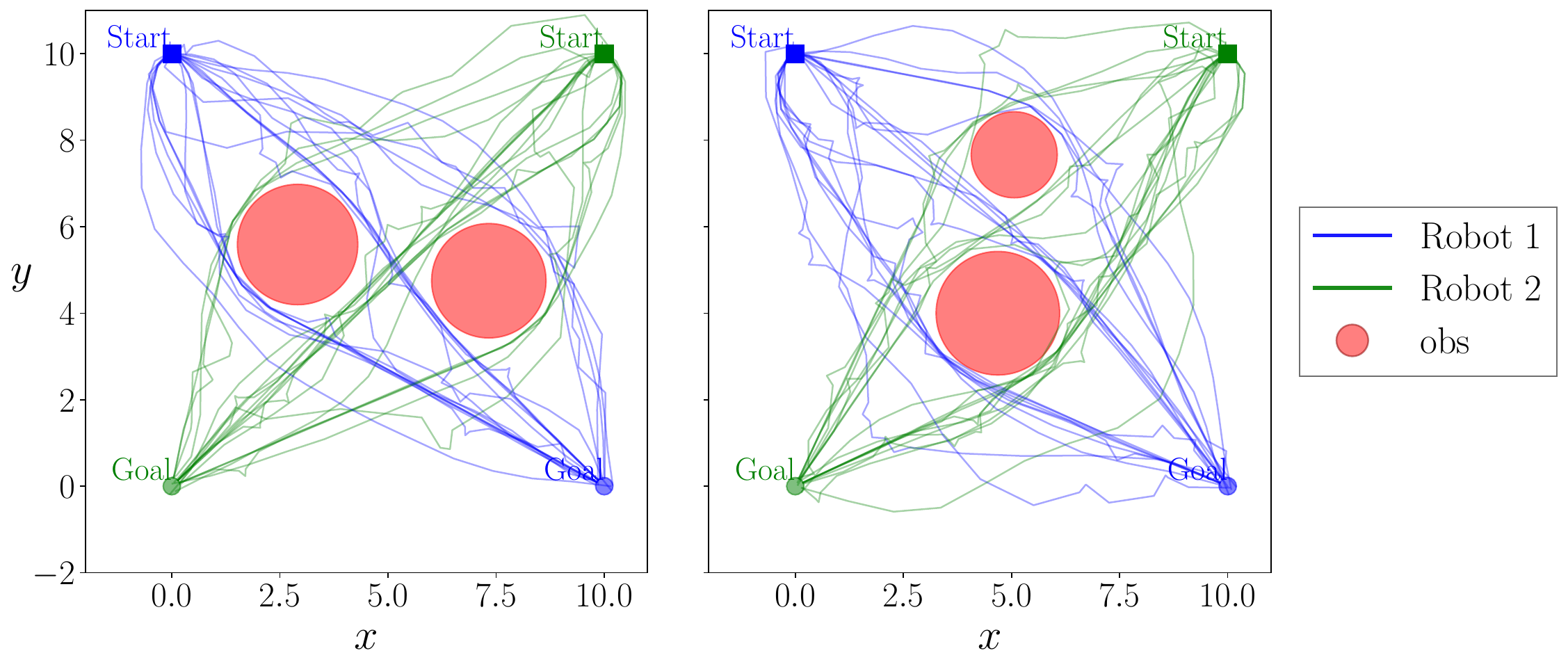}
\caption{
Diverse locally optimal trajectories generated by \methodname for two unseen test configurations in the two-robot navigation example.
}
\label{fig: robot, diverse solution}
\end{figure}

\begin{table}[t]
    \caption{$k$-neighborhood and solving time statistics over locally optimal runs, and the locally optimal ratios over all sampled initial guesses in the two-robot navigation example.
    All samples are used as initial guesses to warm-start the SNOPT solver.
    Higher cumulative distribution values ($\uparrow$) and lower $k$-neighborhood statistics ($\downarrow$) correspond to closer to convergence.
    }
    \label{tab:robot_k_neighbor_solving_time}
    \begin{tabular}{lccccccc}
    \toprule
    Method &
    \multicolumn{3}{c}{$k$-neighborhood Stats. ($\downarrow$)} &
    \multicolumn{3}{c}{Solving Time ($\downarrow$)} &
    Optimal Ratio \% ($\uparrow$) \\
    \cmidrule(lr){2-4}\cmidrule(lr){5-7}
    & Mean & Std & Median & Mean & Std & Median & \\
    \midrule
    Uniform       & 342.17 & 235.39 & 281 & 48.07 & 33.77 & 39.12 & 71.5 \\
    DiffuSolve    & 160.65 & 177.95 & 94  & 22.72 & 26.15 & 12.54 & 84.2 \\
    DiffuSolve-k  & 170.26 & 181.34 & 95  & 24.42 & 26.99 & 13.28 & 84.7 \\
    NS-only       & 136.89 & 153.89 & 76  & 19.55 & 22.68 & 10.66 & 84.0 \\
    GLENS         & \textbf{134.76} & 161.39 & \textbf{71}  
                  & \textbf{19.06} & 23.42 & \textbf{9.76}  
                  & \textbf{85.5} \\
    \bottomrule
    \end{tabular}
\end{table}

At test time, we sample $20$ unseen parameters and generate $50$ samples from each method per parameter instance to warm-start SNOPT.
Figure~\ref{fig: robot, k neighbor cdf} plots the cumulative distribution of the $k$-neighborhood reached by these initial guesses; steeper curves indicate that the corresponding method generates samples closer to the converged optima.
Samples generated by \methodname converge significantly faster than those from all baselines.
This is consistent with Table~\ref{tab:robot_k_neighbor_solving_time}, which summarizes the $k$-neighborhood and solving time statistics over all locally optimal runs.
Compared with the previous examples, the $k$-neighborhood statistics exhibit higher variance, since some instances require up to $1000$ solver iterations to converge; the median is therefore substantially lower than the mean across all methods.
Despite this variability, \methodname achieves the lowest mean and median $k$-neighborhood values, and the highest proportion of samples that reach locally optimal solutions.

We further analyze the computational cost of each method.
For the learning-based methods, diffusion-model inference runs on an NVIDIA L40 GPU and generates $1000$ samples in under one second, so the initial-guess generation time is negligible compared with the overall solving time.
Each initial guess is then used to warm-start SNOPT on a 2.0\,GHz Intel Sapphire Rapids CPU with four cores.
The resulting solving-time statistics are reported in Table~\ref{tab:robot_k_neighbor_solving_time}: initial guesses produced by \methodname require the least time to converge to local optima, consistent with their smaller $k$-neighborhoods.

In addition to producing high-quality initial guesses, \methodname preserves
the multimodality of the solution landscape.
Figure~\ref{fig: robot, diverse solution} illustrates diverse locally optimal trajectories obtained for two unseen test parameter instances, after warm-starting SNOPT with samples generated by \methodname.
Under different obstacle configurations, \methodname discovers multiple distinct routes for the robots to reach their respective goals.

\section{Conclusions}\label{sec: conclusion}

We proposed \methodname, a data-driven and data-efficient global search method for generating high-quality initial guesses in non-convex continuous optimization.
Our approach addresses the data scarcity challenge in existing learning-based methods by using intermediate solver iterates, i.e., the $k$-neighborhood dataset, as free data augmentation.
\methodname combines two complementary models to exploit the $k$-neighborhood dataset: a \neighbormodel that uses diffusion to learn the neighborhood structure around optima, and a \solvermodel that learns the solver's refinement behavior; the two are trained separately and are combined at sampling time to guide diffusion samples toward optimal solutions.

Experimental results show that \methodname effectively exploits the structure contained in solver iterates to generate high-quality initial guesses while preserving solution diversity and multimodality.
Across benchmark problems and a two-robot navigation problem, \methodname generates diverse samples and improves solver convergence under different optimization solvers, including gradient descent, L-BFGS-B, and SNOPT.
The hyperparameter analysis further provides practical guidance for applying \methodname, showing how neighborhood size, model capacity, and guidance strength affect performance.

Future work includes exploring richer conditioning strategies, such as incorporating objective values or constraint violation as additional conditional inputs.
Another exciting direction is to leverage online solver data for adaptation or fine-tuning as new problem instances are encountered.

\FloatBarrier
\backmatter

\bmhead{Acknowledgements}

The experiments in this work were conducted using computational resources supported by Princeton Research Computing, a consortium including the Princeton Institute for Computational Science and Engineering (PICSciE) and the Office of Information Technology's High Performance Computing Center and Visualization Laboratory at Princeton University.

\section*{Declarations}

\begin{itemize}
\item \textbf{Funding.} This work was partially supported by the Princeton Laboratory for Artificial Intelligence's [PLI/AI\textsuperscript{2}/NAM] initiative.
\item \textbf{Competing interests.} 
The authors have no competing interests to declare that are relevant to the content of this article.
\item \textbf{Data availability.} The datasets generated and used during the current study are available at \url{https://huggingface.co/datasets/Anjian/GLENS/tree/main}.
\item \textbf{Code availability.} The code used in this study is available at \url{https://github.com/anjianli21/GLENS}.
\item \textbf{Author contributions.} 
Anjian Li contributed to conceptualization, data curation, investigation, methodology, software, validation, visualization, writing of the original draft, and writing, review, and editing. 
Bartolomeo Stellato contributed to funding acquisition, supervision, and writing, review, and editing of the manuscript. 
Ryne Beeson contributed to conceptualization, funding acquisition, methodology, project administration, resources, supervision, and writing, review, and editing of the manuscript. 
\end{itemize}

\bibliography{Manuscript}

\end{document}